\pdfoutput=1
\PassOptionsToPackage{table}{xcolor}
\documentclass{article} %
\usepackage{iclr2027_conference,times}
\iclrfinalcopy %

\usepackage{amsmath,amsfonts,bm}

\def\eqref#1{equation~\ref{#1}}

\def\1{\bm{1}}

\DeclareMathAlphabet{\mathsfit}{\encodingdefault}{\sfdefault}{m}{sl}
\SetMathAlphabet{\mathsfit}{bold}{\encodingdefault}{\sfdefault}{bx}{n}

\usepackage{hyperref}
\usepackage{url}
\usepackage{booktabs}
\usepackage{graphicx}
\usepackage{amsmath,amssymb}
\usepackage{xcolor}
\usepackage{multirow}
\usepackage{adjustbox}
\usepackage{float}
\usepackage{algorithm}
\usepackage{algpseudocode}
\usepackage{wrapfig}
\usepackage{caption}
\usepackage{tikz}
\usetikzlibrary{shapes.geometric,arrows.meta,positioning,calc}


\newcommand{\method}{PRECOCIAL}
\newcommand{\std}[1]{{\tiny$\pm$#1}}
\definecolor{hl}{HTML}{EEF3FA}
\title{\method{}: Pretraining with Code Curriculums via Iterative Search for Accelerated Learning}

\author{Edward Gu$^{1}$ \quad Ali Payani$^{2}$ \quad Kevin Ellis$^{1}$ \\
$^{1}$Cornell University, $^{2}$Cisco Research \\
\texttt{elg227@cornell.edu}, \texttt{apayani@cisco.com}, \texttt{kellis@cornell.edu}}

\begin{document}

\maketitle
\lhead{Preprint} %

\begin{abstract}
Fully synthetic pretraining data generated by symbolic programs can bootstrap downstream training on naturalistic tasks, such as language modeling, but the generating programs have so far been designed by hand. We introduce \method{}, a method for automatically optimizing programs that generate synthetic pretraining data. Starting from an initial generator program, an LLM iteratively rewrites the program, and each rewrite is scored by pretraining a randomly-initialized transformer on its data and fine-tuning it on the downstream task. Because every score requires training, \method{} keeps a single network for the entire search, continuing its pretraining for each accepted program instead of training a new network per candidate, making it significantly more efficient. On four language and code datasets, at a fixed fine-tuning budget, optimized programs reach lower perplexity than hand-designed curricula, published procedural generators, random program mutation and the program the LLM writes in the first iteration of the search, and reach the same perplexity as the hand-designed curricula with 9--33\% fewer fine-tuning steps. \method{} also ends with lower perplexity than FunSearch and AlphaEvolve, which train a new network for every candidate, at the same number of evaluations and a lower cost. We also observe that an optimized program generator from one domain transfers to datasets in other domains. Additionally, optimized programs allow for better downstream performance on four different reasoning and problem-solving tasks.
\end{abstract}

\section{Introduction}
Large Language Models (LLMs) begin as randomly-initialized transformer networks before pretraining on masses of human data. Must training begin as a randomly-initialized blank slate? Biological learning suggests otherwise. From birth, wildebeests can run and spiders can weave webs, drawing only on their genetic endowment: a compact discrete code (DNA) prewires useful behaviors and biases. Analogously, here we set out to discover compact symbolic programs which prime a randomly-initialized transformer with useful inductive biases, which we call \emph{pretraining programs}.

\begin{figure}[t]
  \centering
  \includegraphics[width=\textwidth]{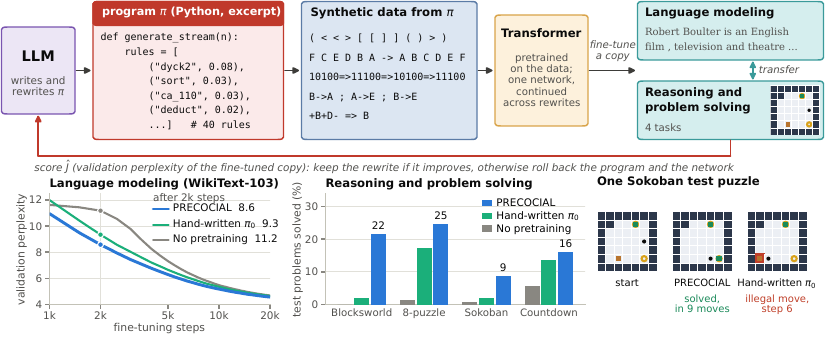}
  \caption{\method{} at a glance. \emph{Top}: an LLM rewrites a program $\pi$ (here, part of the program kept for
  WikiText-103) whose data pretrains one transformer, continued across rewrites; a fine-tuned copy scores each
  rewrite. Programs transfer between tasks (\S\ref{sec:exp-transfer}). \emph{Bottom}, from the runs of
  Tables~\ref{tab:lang} and~\ref{tab:planning}: WikiText-103 validation perplexity by fine-tuning budget, solve
  rates, and the Sokoban puzzle of Figure~\ref{fig:rollouts}.}
  \label{fig:concept}
\end{figure}

Previous work explores two families of pretraining programs: procedures which directly generate weights and biases \citep{stanley2009hypercube,barabasi2023complex}, and programs which generate synthetic data
\citep{zhang2025edge,jiang2026procedural} for ``pre-pretraining.'' Both families could be fruitful, but synthetic data is presently more practical and better-aligned with modern LLM pipelines, and so we adopt it here (Figure~\ref{fig:concept}). Recent work suggests that such pretraining programs can improve both the sample and computational complexity of downstream learning: first training on cellular automata, formal languages, or logical inferences then accelerates downstream learning in natural language, visual reasoning, program source code, and more \citep{zhang2025edge,hu2025circuits,jiang2026procedural,wu2021lime,lee2026nca}.
In all of this work the program is designed by hand. This prior literature naturally suggests our core research question: how could a pretraining program be learned, synthesized, evolved, or automatically improved? This question also has immediate implications for synthetic training data more broadly, as synthetic post-training has become an important way of teaching language models to reason \citep{trinh2024alphageometry,zhao2025absolute}.

To learn a pretraining program, we assume a collection of downstream datasets, such as natural language corpora or reasoning problems. These downstream tasks define an objective for pretraining programs, because we can measure how well a program's synthetic data equips the learner to tackle downstream data. Optimizing this objective is challenging for two reasons: the program search space is discrete, and scoring a single program is expensive, since it requires pretraining a transformer on the program's data and then fine-tuning it on the downstream task. LLM-guided program search such as FunSearch and AlphaEvolve \citep{romeraparedes2024funsearch,novikov2025alphaevolve} score each candidate independently, which is only affordable when the evaluation is cheap. Here, it would mean pretraining a transformer from scratch for every candidate.

To address these challenges we introduce \method{} (PREtraining with COde Curriculums via Iterative search for
Accelerated Learning), a method for synthesizing and optimizing pretraining programs. An LLM rewrites the
program, each rewrite is scored by training on its data, and a rewrite is kept only if it improves the
downstream objective. To make each score cheap, \method{} keeps a \emph{single} network for the whole search:
every accepted program continues pretraining the network where the previous one left off, while a rejected
program rolls back the checkpoint (Section~\ref{sec:continual}). Our goal is \emph{not} to show that pretraining programs
can help downstream tasks, which is already well-established. Instead, our primary goal is to evaluate whether
such pretraining programs can be automatically optimized for a dataset of interest, and whether such
optimization \emph{transfers} across datasets. This paper makes the following contributions:
\begin{itemize}\setlength{\itemsep}{1pt}\setlength{\parskip}{0pt}
  \item \textbf{Algorithm.} \method{}: LLM-guided search over pretraining programs whose fitness is the
  downstream performance of a network trained on the program's data (\S\ref{sec:method}).
  \item \textbf{Efficient continual pretraining.} \method{} maintains a single model that is continuously
  pretrained across candidate programs. At a fixed fine-tuning budget this is both cheaper and better than
  training a fresh model per candidate (i.e.\ naive application of FunSearch/AlphaEvolve): \method{} ends with lower perplexity
  than both while using 2.5$\times$ fewer pretraining steps (\S\ref{sec:exp-continual}).
  \item \textbf{Empirical validation.} On four language and code datasets, optimized programs reach lower
  perplexity than every other source of a curriculum we test (hand-written, procedural, random mutation, the
  LLM's first-iteration program) at every fine-tuning budget we measure, and they transfer to unseen
  datasets. At a 10k-step fine-tuning budget, \method{} wins on all four reasoning and problem-solving tasks, and a Blocksworld search that never sees the task does as well as one that does (\S\ref{sec:exp}).
\end{itemize}

\section{Background and Related Work}
\textbf{Structural synthetic pretraining.} A growing body of work shows that pretraining on structured
synthetic data teaches neural networks inductive biases that transfer to natural tasks. Cellular automata at
the edge of chaos yield models with strong compositional structure \citep{zhang2025edge}; neural cellular
automata seed persistent memory circuits \citep{lee2026nca}; and the phenomenon extends to vision
transformers \citep{shinnick2025see}. On the formal language side, Dyck languages and context-free grammars
teach bracket tracking and hierarchical composition \citep{hu2025circuits,papadimitriou2020music}, while
LIME's synthetic inference tasks aid compositional generalisation \citep{wu2021lime}. A common theme is that
\emph{abstract structural patterns}, not semantic content, drive transfer, and the benefit is usually one of
efficiency: pre-pretraining on formal languages lets a 1B-parameter model reach the final loss of a model without it with 33\% fewer
tokens \citep{hu2025circuits}, and procedural data lets models reach the same loss with 55--86\% of the original
data \citep{jiang2026procedural}. This is also mostly what pretraining on real data buys: detectors trained from
scratch on COCO match ImageNet-pretrained ones once given a longer schedule \citep{he2019rethinking}, and
pretraining helps most, relative to the fine-tuning data, when that data is scarce \citep{hernandez2021transfer}.
All of these generators are fixed by their designers. We compare against the procedural generators of
\citet{jiang2026procedural} directly, and report our gains in the same currency of steps saved.

\textbf{LLM-guided program search.} FunSearch \citep{romeraparedes2024funsearch} and AlphaEvolve
\citep{novikov2025alphaevolve} use LLMs as mutation operators over programs, maintaining populations selected
by black-box fitness, and are applied where a candidate is scored by running it. \method{} uses the same
LLM-as-mutator paradigm, but its fitness requires training a network. We therefore search over about ten
candidates rather than thousands, keep one continually pretrained network across them, and compare against
both methods at the same budget.

\textbf{Learning to generate training data.} Generative Teaching Networks \citep{such2020gtn} learn a neural
data generator by meta-gradients; Montessori-Instruct \citep{li2025montessori} trains a teacher LLM on
instruction data using a student's data influence; DataEvolve \citep{mi2026dataevolve} evolves curation
strategies for natural pretraining data with trained models as fitness; CADEvolve \citep{elistratov2026cadevolve}
evolves data-generating programs selected by validity checks; and Eurekaverse \citep{liang2024eurekaverse} has
an LLM write training environments scored by a reinforcement learner. Most closely related, self-play pretraining
\citep{cowsik2026selfplay} trains a transformer by reinforcement learning to write programs for a universal Turing
machine whose outputs lie at the frontier of what a learner can learn, with no natural data. \method{} differs in
two ways: it optimizes for downstream tasks, whereas self-play is unsupervised, and it yields one compact program
that describes the pretraining distribution, whereas in self-play that distribution is defined by the weights of
the generator network. Our work is also
related to meta-learning \citep{finn2017maml} and learned data augmentation \citep{ho2019pba}. We are not aware
of prior work that searches over programs that generate pre-pretraining data with the downstream performance of
a network trained on that data as the fitness.

\section{Methods}\label{sec:method}
Given a downstream dataset $\mathcal{D}_\text{down}$, our goal is to arrive upon a data generating program
$\pi$ whose outputs prime a transformer to do well on that dataset, and potentially generalise to unseen
downstream tasks as well. Each downstream dataset comprises train/val/test splits and a scoring metric,
$(\mathcal{D}_\text{train}, \mathcal{D}_\text{val}, \mathcal{D}_\text{test}, \mathcal{L})$. We view each data
generating program $\pi$ as defining a distribution over token sequences, and also view a downstream training
split as a distribution over token sequences. We write $\theta(\mathcal{D}_1, \mathcal{D}_2, \dots)$ for the
parameters of a transformer which is first trained on $\mathcal{D}_1$, then $\mathcal{D}_2$, etc. Using this
notation, we seek a program $\pi$ optimising
\begin{equation}
  \hat{J}(\pi) = \mathcal{L}(\mathcal{D}_\text{val}, \theta(\pi, \mathcal{D}_\text{train})),\label{eq:obj}
\end{equation}
where $\hat{J}$ approximates the true objective (which uses $\mathcal{D}_\text{test}$) on the validation data,
ensuring that test data is never seen during program search. We represent programs $\pi$ in Python. Our
transformers are standard autoregressive models, except that we use byte-level tokenisation: tokenisers are
constructed from massive human data, and a key part of our problem statement is to arrive upon a pretrained
state \emph{without} massive human data.

Figure~\ref{fig:concept} summarises the method; Figure~\ref{fig:method} and Algorithm~\ref{alg:precocial} in
Appendix~\ref{app:alg} give it with the notation of this section and as pseudocode.

\textbf{Initial program.} An important choice is the starting point of the search. We construct a hand-written
program $\pi_0$ by taking the union of existing synthetic pretraining paradigms from the recent literature
(Table~\ref{tab:init}); it serves as a strong baseline. Programs follow a fixed generator contract
(Appendix~\ref{app:contract}): each emits streams of abstract ASCII symbols, tagged with the rule that produced
them; English words are rejected by a validation step.

\begin{table}[t]
  \caption{Elements of the hand-written program $\pi_0$ (28 rules).}
  \label{tab:init}
  \centering\scriptsize\setlength{\tabcolsep}{4pt}
  \begin{adjustbox}{max width=\linewidth}\begin{tabular}{@{}lll@{}}
    \toprule
    Tier & Rules & Literature source / rationale \\
    \midrule
    1. Hierarchical & \texttt{dyck1/2, dyck\_shuf, nest1--3} & Bracket tracking, recursive composition \citep{hu2025circuits} \\
    2. Cellular automata & \texttt{ca\_rule} & Edge-of-chaos rules maximise transfer \citep{zhang2025edge} \\
    3. Sequence ops & \texttt{sort, reverse, union, inter, copy, dedup} & Set and sequence primitives \citep{wu2021lime} \\
    4. Reasoning & \texttt{deduct, induct, abduct} & Three modes of primitive inference \citep{wu2021lime} \\
    5. SCAN-style & \texttt{twice, thrice, after, around} & Compositional commands \citep{lake2018scan} \\
    6. Procedural/CFG & \texttt{cfg, arith, mod, swap, pair} & CFG derivations and arithmetic \citep{jiang2026procedural} \\
    7. State tracking & \texttt{stack, fsm, counter} & Push/pop and FSM; persistent memory \citep{lee2026nca} \\
    \bottomrule
  \end{tabular}\end{adjustbox}
\end{table}

\subsection{LLM-guided search for pretraining programs}\label{sec:search}
Following AlphaEvolve and others, \method{} generates variants of a program via LLM prompting and keeps the
best under $\hat{J}$. In the first iteration of the search, the LLM writes a program from the generator
contract, which lists the rule families of Table~\ref{tab:init}, and a sample of downstream data. We call this the
\emph{first-iteration program}; it reflects the LLM's prior, before  feedback from training, and
80--89\% of the data in these programs comes from the rules of $\pi_0$. At each subsequent round the LLM receives the current
accepted program, its objective value and the best so far, the validation curves of the last two candidates,
300 bytes of downstream training data, and returns a complete rewritten program: it may reweight, remove, rewrite or invent rules freely
(Appendix~\ref{app:prompt}). The rewrite is scored by Eq.~\ref{eq:obj}. It is accepted if
$\hat{J}(\pi_k) \le \hat{J}^*$, the best value so far; otherwise the search reverts to the last accepted
program.

\subsection{Continual pretraining across candidates}\label{sec:continual}
Our objective function is very expensive to evaluate, requiring both pretraining and fine-tuning. This is in
stark contrast to most program synthesis problems, where evaluation is so fast that even guess-and-check
strategies are often viable \citep{li2022alphacode,schkufza2013stoke}. After running this expensive
evaluation, a naive approach would throw away hard-won pretrained parameters before generating and evaluating
another program, as FunSearch and AlphaEvolve do when each candidate is scored independently. Instead, \method{} \emph{continues} pretraining the same model from the
last accepted checkpoint with each new program: the first program pretrains a randomly initialized network for
$T_0$ steps, and every later candidate continues it for $T_c < T_0$ steps. A rejected candidate's steps are
discarded, so the network always reflects the accepted programs only. This can be viewed as a form of continual
learning, but also of curriculum learning: the system gradually enriches the pretraining program, with each
program resuming where the previous one left off. With $T_0{=}15$k, $T_c{=}5$k and ten candidates, a search
performs 60k pretraining steps, against 150k for a fresh 15k-step network per candidate.

\textbf{Comparison to FunSearch and AlphaEvolve.} Both maintain a population of programs from which parents
are sampled; \method{} is a single chain. We implement both as published, with a fresh network trained for every
candidate, at the same budget of ten evaluations (Appendix~\ref{app:setup}).

\section{Experiments}\label{sec:exp}
\subsection{Setup}\label{sec:setup}
\textbf{Downstream datasets.} \emph{Language and code:} TinyStories \citep{eldan2023tinystories} (short
narrative stories), WikiText-103 \citep{merity2017pointer} (encyclopedia articles), CodeParrot
\citep{codeparrot} (Python source code) and CodeSearchNet \citep{husain2019codesearchnet} (Python functions),
with 10k training documents each (TinyStories: 7{,}818 stories of at most 512 characters), byte-encoded as raw
text, and test sets from held-out data (Appendix~\ref{app:data}). \emph{Reasoning and problem solving:} Blocksworld with 9 to 12 blocks, a standard benchmark of
reasoning about actions for language models \citep{valmeekam2023planbench} (input: a PDDL problem; output: a plan, solved if
replaying it in a simulator reaches the goal); the 8-puzzle (states 10--20 moves from the goal); Sokoban
($7{\times}7$, two boxes); and Countdown with four numbers \citep{gandhi2024sos}. Each has 5{,}000 training problems
(Blocksworld 10{,}000) and is scored by an exact checker on 500 test problems (Blocksworld 200), with greedy
decoding.

\textbf{Model training.} We use small 0.69M-parameter GPT-2 models (3 layers, 4 heads, 128-dimensional
embeddings) with byte-level tokenization and a 512-byte context (1{,}024 for Blocksworld), optimised with AdamW. A search runs $K{=}10$ rounds with $T_0{=}15$k and $T_c{=}5$k
steps on 2{,}000 synthetic streams per round. Each candidate is fine-tuned for up to 10k steps with early
stopping %
and $\hat{J}$ is validation perplexity (on the answer tokens,
for the puzzles). The LLM is Claude Sonnet~4 \citep{anthropic2025claude4}. The 10k-step budget and the ten
rounds were set once and used unchanged for every dataset, task and baseline (Appendix~\ref{app:setup}).

\textbf{Baselines.} All share the architecture and training budget. \emph{Hand-written} trains on $\pi_0$
(Table~\ref{tab:init}). \emph{Procedural} trains on the published generators of \citet{jiang2026procedural}.
\emph{First iteration} trains on the first-iteration program (Section~\ref{sec:search}), the LLM's program before any search, to tell whether the LLM or the optimization drives the improvement. \emph{Random mutation} runs the same loop with the LLM replaced by random edits of that
program (reweight all rules, remove one, or duplicate one). \emph{No pretraining} fine-tunes a randomly
initialized network.

\textbf{Matched pretraining.} Comparing methods at the end of optimization would be confounded: a rejected
candidate is rolled back, so runs accumulate different amounts of pretraining (15k to 55k steps across our runs). We therefore
compare curricula under \emph{matched pretraining}: the best program of \method{} and of every baseline pretrains a fresh network for exactly
60k steps, then gets the same fine-tuning and test: test perplexity after a 10k-step fine-tune for language, the
exact checker after a 10k-step fine-tune for puzzles (Appendix~\ref{app:setup}).
Section~\ref{sec:exp-continual}, which compares search procedures, evaluates the network each search produces. We report mean\,$\pm$\,SD over runs and two-sided $t$-tests on paired differences.

\subsection{Language and code: search against every other source of a curriculum}\label{sec:exp-lang}
\begin{table}[t]
  \caption{Language and code under matched pretraining: test perplexity after a 10k-step fine-tune (lower is
  better), mean\,$\pm$\,SD over 9 runs. %
  Last column: difference from
  \method{}  over the four datasets (positive: \method{} lower; 36 paired runs), $p$ in parentheses; bold:
  $p<.05$. Rounds of one search:
  the first iteration is round 0, \method{} the program after round 9, and the bottom row the program
  after round 5.}
  \label{tab:lang}
  \centering\small\setlength{\tabcolsep}{3.3pt}
  \begin{adjustbox}{max width=\linewidth}\begin{tabular}{@{}lccccc@{}}
    \toprule
    Pretraining & WikiText-103 & TinyStories & CodeParrot & CodeSearchNet & $\Delta$ vs \method{} \\
    \midrule
    \rowcolor{hl}\method{} & 5.284\std{.126} & 2.810\std{.081} & 3.207\std{.092} & 3.996\std{.177} & -- \\
First iteration & 5.455\std{.044} & 2.875\std{.018} & 3.293\std{.032} & 4.295\std{.453} & \textbf{$+$0.155} {\scriptsize($<$.001)} \\
Random mutation & 5.471\std{.041} & 2.877\std{.032} & 3.306\std{.042} & 4.205\std{.117} & \textbf{$+$0.140} {\scriptsize($<$.001)} \\
Procedural & 5.688\std{.074} & 2.942\std{.025} & 3.388\std{.051} & 4.766\std{.581} & \textbf{$+$0.371} {\scriptsize($<$.001)} \\
Hand-written ($\pi_0$) & 5.466\std{.033} & 2.887\std{.028} & 3.298\std{.024} & 4.495\std{.546} & \textbf{$+$0.212} {\scriptsize($<$.001)} \\
No pretraining & 5.595\std{.045} & 2.932\std{.061} & 3.204\std{.015} & 5.511\std{.416} & \textbf{$+$0.486} {\scriptsize($<$.001)} \\
\midrule
\textit{\method{} after round 5} & 5.296\std{.106} & 2.814\std{.063} & 3.227\std{.133} & 4.077\std{.131} & $+$0.029 {\scriptsize(.069)} \\
 
    \bottomrule
  \end{tabular}\end{adjustbox}
\end{table}

\begin{figure}[t]
  \centering
  \includegraphics[width=\textwidth]{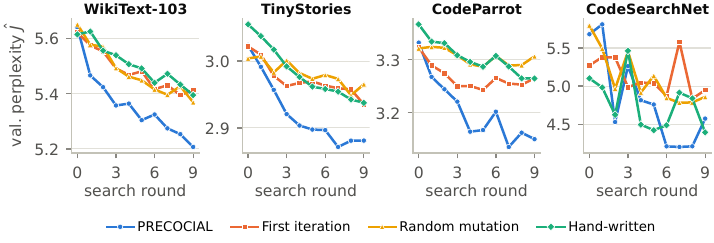}
  \caption{The search objective over rounds: validation perplexity $\hat{J}$ of each round's candidate, mean over
  3 seeds. \method{} improves over the rounds and separates from the baselines on three of the
  four datasets; on CodeSearchNet every method's trajectory is noisy, and the kept program is the best round, not the
  last.}
  \label{fig:dynamics}
\end{figure}

Table~\ref{tab:lang} compares \method{} with every other way of obtaining a curriculum. At the 10k-step
budget \method{} has the lowest test perplexity of any curriculum on all four datasets, and it is level with no
pretraining only on CodeParrot (3.207 against 3.204). Pooled over the four datasets, it improves on the first-iteration
program by 0.155 (lower in 30 of 36 paired runs), on random mutation by 0.140 (32 of 36), on the
hand-written program by 0.212 (31 of 36) and on procedural pretraining by 0.371 (36 of 36), all with $p{<}.001$.
Since the first-iteration and random-mutation baselines start from the same program as the search, the gain over them is due to
the LLM-guided search loop. Program quality  improves  over the rounds: under matched pretraining the
program kept after round 5 already beats the first-iteration program (round 0) by 0.126 ($p{=}.005$; 28 of 36), and the last four
rewrites add a further 0.029 ($p{=}.07$). On every dataset round 9 beats round 5, which beats round 0. Without matched
pretraining, at the end of the ten-round loop, \method{} is again lowest on every dataset (Appendix~\ref{app:loop}).
Figure~\ref{fig:dynamics} plots the search itself.

\textbf{Across fine-tuning budgets} (Figure~\ref{fig:budget} in Appendix~\ref{app:tables}), \method{} is ahead of all four alternative
curricula at every budget from 2k to 20k steps, significantly so against every baseline at every budget
(Table~\ref{tab:budget-matched}). The lead is largest early, 0.78--1.55 perplexity at 2k steps, and narrows to
0.06--0.14 by 20k steps. Searching therefore buys downstream
learning speed. Measured in fine-tuning steps, as \citet{hu2025circuits} and \citet{jiang2026procedural} do,
\method{} reaches its 2k-step perplexity with 57\%, 67\% and 80\% of the steps that no pretraining, procedural
pretraining and the hand-written program need, and its 10k-step perplexity with 83\%, 80\% and 89\% (medians over
36 runs; Table~\ref{tab:savings}).

\subsection{\method{} against FunSearch and AlphaEvolve}\label{sec:exp-continual}
Naive application of FunSearch and AlphaEvolve to this setting would score every candidate on a network pretrained from scratch for it, whereas \method{}
continues one network across the whole search (Section~\ref{sec:continual}). We run all three at the same budget
of ten evaluations on WikiText-103 and CodeSearchNet, three seeds each, with everything else identical, including
the GPU type (Table~\ref{tab:continual}).
Table~\ref{tab:continual} shows the result. \method{} ends with lower test perplexity than both methods on both
datasets: averaged over the two, 0.231 below FunSearch ($p{=}.15$; lower in 4 of 6 paired runs) and 0.296 below
AlphaEvolve ($p{=}.037$; 5 of 6). It is also cheaper. FunSearch and AlphaEvolve pretrain each of the ten candidates
from random initialisation for $T_0{=}15$k steps, 150k steps in total, whereas \method{} pretrains one network for
15k steps and continues it by $T_c{=}5$k for each later candidate, 60k in total. Fine-tuning is identical, so
\method{} uses 19.5\% and 17.1\% less GPU time per search, for about the same LLM bill; fine-tuning the candidates
is the largest part of every search, so the saving is modest here, and it grows with the number of candidates and
with model size (Appendix~\ref{app:cost}). The continued network is what makes the difference: run with a fresh
network for every candidate, \method{} itself ends 0.244 perplexity higher ($p{=}.055$; 5 of 6). Part of the gain
is that the continued network accumulates more pretraining (up to 60k steps, against 15k for each fresh network),
a benefit obtained at lower cost.

\begin{table}[t]
  \caption{\method{} against FunSearch and AlphaEvolve baselines which pretrain a fresh network for every
  program, at the same budget of ten evaluations. Test perplexity after a 10k-step fine-tune (lower is better),
  mean\,$\pm$\,SD over 3 seeds. $\Delta$: difference from \method{} pooled over the two datasets (positive:
  \method{} lower; 6 paired runs), $p$ in parentheses; bold: $p<.05$. Last two columns: synthetic pretraining
  steps and GPU-hours of one search (ten evaluations, RTX A6000). Bottom row: \method{} itself with a fresh network
  per candidate. Every row tests the network its own search produces (no matched pretraining); the \method{} row
  is an independent repeat of the ten-round loop of Table~\ref{tab:lang-loop}.}
  \label{tab:continual}
  \centering\footnotesize\setlength{\tabcolsep}{3.5pt}
  \begin{adjustbox}{max width=\linewidth}\begin{tabular}{@{}lccccc@{}}
    \toprule
    Search & WikiText-103 & CodeSearchNet & $\Delta$ vs \method{} & Pretraining steps & GPU-h \\
    \midrule
    \rowcolor{hl}\method{} & 5.151\std{.201} & 4.167\std{.149} & -- & 60,000 & 0.71 \\
FunSearch & 5.553\std{.041} & 4.227\std{.360} & $+$0.231 {\scriptsize(.152)} & 150,000 & 0.88 \\
AlphaEvolve & 5.496\std{.127} & 4.414\std{.167} & \textbf{$+$0.296} {\scriptsize(.037)} & 150,000 & 0.85 \\
\midrule
\method{} with a fresh network per candidate & 5.507\std{.128} & 4.299\std{.399} & $+$0.244 {\scriptsize(.055)} & 150,000 & 0.85 \\
 
    \bottomrule
  \end{tabular}\end{adjustbox}
\end{table}

\subsection{Transfer to unseen datasets}\label{sec:exp-transfer}
Does a program optimized for one dataset help others? We take the network pretrained on the WikiText-103
program of \method{} and of each baseline (matched pretraining) and fine-tune it on TinyStories, CodeParrot and CodeSearchNet for
10k steps, with no further search (Table~\ref{tab:transfer}).

\begin{table}[t]
  \caption{Transfer: networks pretrained on programs obtained for WikiText-103, fine-tuned for 10k steps on a
  different dataset with no further search (matched pretraining). Test perplexity, mean\,$\pm$\,SD over 9 runs
  (3 searches $\times$ 3 training seeds); last column pooled over the three targets (27 paired runs; positive:
  \method{} lower). Bottom row: the program
  \method{} found when searching on the target itself (Table~\ref{tab:lang}).}
  \label{tab:transfer}
  \centering\small\setlength{\tabcolsep}{3pt}
  \begin{adjustbox}{max width=\linewidth}\begin{tabular}{@{}lcccc@{}}
    \toprule
    WikiText-103 program & TinyStories & CodeParrot & CodeSearchNet & $\Delta$ vs \method{} \\
    \midrule
    \rowcolor{hl}\method{} & 2.804\std{.070} & 3.215\std{.067} & 4.279\std{.508} & -- \\
First iteration & 2.897\std{.031} & 3.322\std{.024} & 4.363\std{.227} & $+$0.095 {\scriptsize(.058)} \\
Random mutation & 2.888\std{.038} & 3.320\std{.039} & 4.655\std{.665} & $+$0.188 {\scriptsize(.120)} \\
Procedural & 2.944\std{.014} & 3.406\std{.022} & 4.620\std{.424} & \textbf{$+$0.224} {\scriptsize(.011)} \\
Hand-written ($\pi_0$) & 2.868\std{.016} & 3.309\std{.017} & 4.228\std{.032} & $+$0.035 {\scriptsize(.550)} \\
No pretraining & 2.915\std{.058} & 3.204\std{.017} & 5.632\std{.327} & \textbf{$+$0.484} {\scriptsize(.002)} \\
\midrule
\textit{Optimized on the target} & 2.810\std{.081} & 3.207\std{.092} & 3.996\std{.177} & \\
 
    \bottomrule
  \end{tabular}\end{adjustbox}
\end{table}

The program optimized for WikiText-103 transfers well to TinyStories and CodeParrot, where it is the best
curriculum and within 0.008 perplexity of a program optimized on the target itself, and less well to CodeSearchNet. Pooled over the three targets it beats procedural pretraining ($+0.224$, $p{=}.011$; 25 of 27 paired runs)
and no pretraining ($+0.484$, $p{=}.002$), is ahead of the first-iteration program ($+0.095$, $p{=}.058$) and random mutation
($+0.188$, $p{=}.12$), and is level with the hand-written program ($+0.035$, $p{=}.55$).  Transfer even crosses domains (Appendix~\ref{app:xdomain}). The programs
optimized on Blocksworld and Sokoban (Section~\ref{sec:exp-planning}) do as well on the four language datasets as
the programs optimized on those datasets themselves (within 0.017 and 0.003
perplexity, $p{\ge}.69$), and in the other direction the WikiText-103 program solves more problems than the
hand-written program on all four reasoning tasks, though no single difference is significant. These tasks and language have nothing in common at the task level, so this only makes sense if what transfers is the same structural skill (nesting, sequence operations, state tracking) rather than anything task-specific.

\subsection{Reasoning and problem solving from scratch}\label{sec:exp-planning}
Do optimized programs help a small model learn tasks that require multi-step reasoning and problem solving? Each task has three optimized programs (one per search seed), each
trained with several training seeds: three for Blocksworld (9 runs per method) and two for the others (6 runs per
method).

\begin{table}[t]
  \caption{Reasoning and problem solving from scratch, matched pretraining (each curriculum pretrains a fresh network
  for 60k steps; every method then gets the same 10k-step fine-tune). Test solve rate by an exact checker, mean\,$\pm$\,SD
  over 9 runs (Blocksworld) or 6 runs (others). Bottom rows: seed-paired difference, $p$ in parentheses; bold:
  $p<.05$. No pretraining: 6 training seeds, not paired.}
  \label{tab:planning}
  \centering\small\setlength{\tabcolsep}{3pt}
  \begin{adjustbox}{max width=\linewidth}\begin{tabular}{@{}lcccc@{}}
    \toprule
    Pretraining & Blocksworld (9--12 blocks) & 8-puzzle & Sokoban & Countdown \\
    \midrule
    \rowcolor{hl}\method{} & 0.216\std{.254} & 0.246\std{.028} & 0.086\std{.030} & 0.161\std{.012} \\
First iteration & 0.053\std{.158} & 0.156\std{.077} & 0.013\std{.006} & 0.156\std{.014} \\
Hand-written ($\pi_0$) & 0.019\std{.040} & 0.172\std{.035} & 0.019\std{.013} & 0.136\std{.019} \\
No pretraining & 0.000\std{.000} & 0.014\std{.011} & 0.007\std{.003} & 0.056\std{.039} \\
\midrule
\method{} $-$ hand-written & \textbf{$+$0.196} {\scriptsize(.033)} & \textbf{$+$0.073} {\scriptsize(.005)} & \textbf{$+$0.067} {\scriptsize(.002)} & \textbf{$+$0.025} {\scriptsize(.025)} \\
\method{} $-$ first iteration & $+$0.163 {\scriptsize(.060)} & \textbf{$+$0.090} {\scriptsize(.036)} & \textbf{$+$0.073} {\scriptsize(.003)} & $+$0.005 {\scriptsize(.104)} \\
 
    \bottomrule
  \end{tabular}\end{adjustbox}
\end{table}

\textbf{Optimized programs beat the hand-written curriculum on all four tasks} at the 10k-step fine-tuning
budget (Table~\ref{tab:planning}):
Sokoban $+0.067$ ($p{=}.002$), the 8-puzzle $+0.073$ ($p{=}.005$), Countdown $+0.025$ ($p{=}.025$) and Blocksworld
$+0.196$ ($p{=}.033$), ahead in every paired run on the first three. Blocksworld is all-or-nothing: across the 33 runs of Table~\ref{tab:planning}
every model that reaches a test perplexity below 1.04 solves some problems and every model above it solves none
(Appendix~\ref{app:planning}), and 6 of 9 runs of the optimized programs land below that line, against 1 of 9 for the first-iteration program
and 2 of 9 for the hand-written program (Figure~\ref{fig:planning} in Appendix~\ref{app:planning}).
At this budget a network with no synthetic pretraining learns almost nothing on the three tasks whose answer is a sequence of moves
(solve rates 0.000 on Blocksworld, 0.014 on the 8-puzzle and 0.007 on Sokoban
\unskip).

\textbf{The search loop matters, not only the LLM's prior.} Every search starts from the first-iteration program, so
comparing against it isolates what the rounds of search add. The optimized programs beat it
significantly on Sokoban ($+0.073$, $p{=}.003$) and the 8-puzzle ($+0.090$, $p{=}.036$), and lead it on Blocksworld
($+0.163$, $p{=}.06$). On Sokoban the first-iteration program is no better than the hand-written program, so the whole gain there
comes from search. Countdown is the exception: its first-iteration program is already close to the optimized program ($+0.005$,
$p{=}.10$). The search adds most where the LLM's first guess at a curriculum is poor, and little where that guess,
like $\pi_0$, already covers the task.

\begin{figure}[t]
  \centering
  \includegraphics[width=\textwidth]{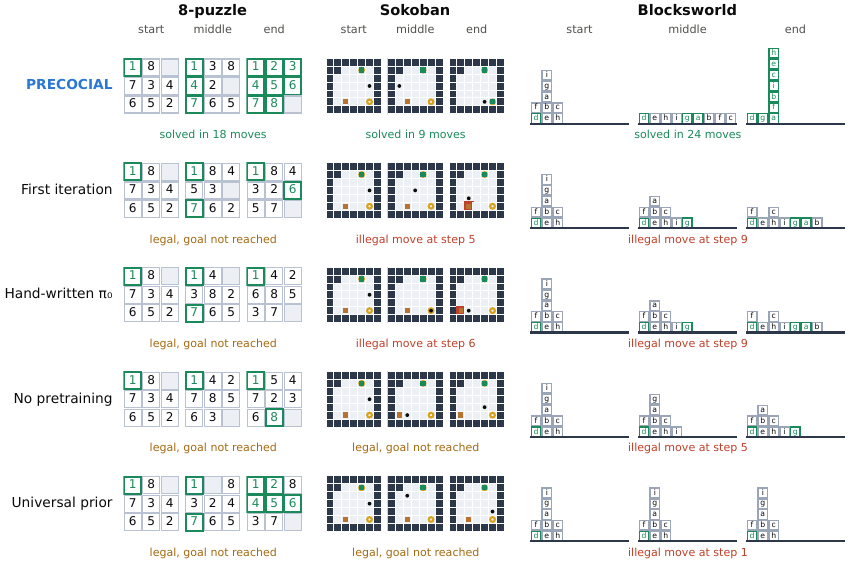}
  \caption{One test puzzle per task, and the start, middle and end of the plan each method's network writes for
  it, replayed by the task's exact checker; below each row, the checker's verdict. Networks trained as in
  Table~\ref{tab:planning} with training seed 64. \emph{Universal prior}: pretrained on the outputs of random
  programs, as in \citet{graumoya2024universal}. Green marks a tile, box or block in its goal position; in Sokoban
  the dot is the player and a red square marks the cell an illegal move tries to enter.}
  \label{fig:rollouts}
\end{figure}

\textbf{What the plans look like.} Figure~\ref{fig:rollouts} shows plans
written by one network of each method. On these puzzles the \method{} network reaches the goal, while every other
network either makes an illegal move or ends its plan short of the goal.
Solve rates are in
Table~\ref{tab:planning}, and Appendix~\ref{app:rollouts} gives the rule and the full plans.

\subsection{What the LLM writes}\label{sec:exp-rules}
Every program is a weighted set of named rules, each generating one kind of sequence. We classify rules by
name into content types (Appendix~\ref{app:rules}) and measure, for each task, the share of synthetic data
coming from rules of the type that matches the task.

\textbf{The LLM writes task-specific rules beyond the human-written $\pi_0$.} For example: \texttt{precond} and
\texttt{state\_trans} for Blocksworld, \texttt{grid\_move} and \texttt{sokoban\_push} for the move puzzles,
\texttt{arith\_precedence} for Countdown, \texttt{func\_def} and \texttt{try\_catch} for code. The first-iteration programs contain almost none of this (0.0--8.4\% of the data;
Table~\ref{tab:rules} in Appendix~\ref{app:rules}); the share grows over the rounds of search (Figure~\ref{fig:rules}) to
3.3--22.8\%.

\textbf{But task-specific content is not what the search selects for.} Comparing, within each task and round,
the candidates the search accepted with those it rejected, accepted candidates carry no more task-matching data
(+2.2 points, stratified permutation test $p{=}.35$, 162 candidates). A direct test agrees. We re-ran the
Blocksworld search with the downstream data hidden from the LLM, which then sees only its programs and their
scores: it wrote no planning rules at all (0 of 298 added rules), yet under matched pretraining its programs
solve as many problems as those of the search that sees the task (0.196 against 0.216, $p{=}.82$, 9 runs each).
Much of what the search keeps is general structure: the 28 generic rules of $\pi_0$ (nesting,
sequence operations and state tracking, among others), which any byte-level sequence model can use, still produce
43--74\% of the kept programs' data, down from 80--89\% in the first-iteration programs. This may explain why programs transfer across domains (Section~\ref{sec:exp-transfer}): the puzzle
programs that transfer best to language, Blocksworld and Sokoban, carry the least task-specific data (3.3\% and
12.3\% of their kept programs, against 22.5\% and 22.8\% for the 8-puzzle and Countdown), though four programs
are too few to be sure.

\section{Discussion}

We set out to ask whether a pretraining program can be optimized automatically, and at a fixed fine-tuning budget
our answer is yes. An LLM that rewrites a generator program, scored by training a small network on the program's
data, finds programs that beat the hand-written curriculum, published procedural generators, random mutation and its
own first-iteration program on four language and code datasets, and that beat the hand-written curriculum on four
reasoning and problem-solving tasks. The programs improve steadily over the rounds of search, and they transfer to datasets they were not
optimized on, in part even across domains.

\textbf{What makes the search work.} Our results point to the trained model rather than the LLM's knowledge of the
task. On Blocksworld a search that never sees the task does as well as one that does, and across tasks the programs
the search accepts carry no more task-specific content than those it rejects. The LLM is still not interchangeable
with random edits: random mutation, which cannot write new rules, falls short of it on language. Our reading is that
the LLM's job is to propose varied, valid programs, including kinds of data the current program lacks, and that
scoring each proposal on a trained network is what decides which of them help. The same view accounts for transfer:
much of what survives is general structure that any sequence model can use.

\textbf{Evaluating candidates when fitness requires training.} Because every score requires training, how
candidates are evaluated matters as much as how they are proposed. Continuing one network across candidates,
instead of training a fresh network for each as FunSearch and AlphaEvolve do, is both cheaper and better at a
fixed budget.

\textbf{When is this useful?} Like other synthetic pretraining papers, the benefit we measure is efficiency: the
optimized programs reach a given perplexity in fewer fine-tuning steps (Table~\ref{tab:savings}), and their lead is
largest with the least downstream data (Appendix~\ref{app:sweeps}). \method{} is therefore aimed at settings where
downstream data is scarce and no more human data is available, since synthetic data costs nothing to sample. And
because programs transfer, a search can be run once and its program reused on related datasets.

\textbf{Connection to epiplexity.} Epiplexity is the structural information that a computationally bounded learner
can extract from data, estimated by the area under its loss curve above the final loss \citep{finzi2026epiplexity}.
A pretraining program is a compact source of it, and \method{} produces new epiplexity at every iteration: each
rewrite gives the continued network new structure to learn, so its pretraining loss rises at 32 of the 36 round
boundaries of the language searches and falls again within every round (Figure~\ref{fig:pretrainloss}), and the
search keeps the rewrites whose new structure helps downstream. Across the 26 programs of this paper, the
epiplexity a fresh 0.69M network extracts goes with lower WikiText-103 perplexity (Spearman $\rho{=}{-}0.64$,
$p{<}.001$) while the final pretraining loss does not, and the optimized programs carry more of it than their
first-iteration programs in all three WikiText-103 searches (Appendix~\ref{app:epiplexity}).

\textbf{Limitations and future work.} Our models are small (0.15M--3.36M parameters, Appendix~\ref{app:sweeps}):
across that range the head start grows with the model but is spent sooner, so the lead at a fixed budget shrinks with
size. Larger models, larger downstream datasets and longer searches were beyond our compute budget. Each task used
three searches of ten evaluations with a single LLM (Claude Sonnet~4), so smaller effects remain unresolved, and the task-hidden test was run on Blocksworld only. Mixing different
datasets together, and mixing the discovered pretraining data into fine-tuning, may improve results further,
especially for already pretrained models such as Llama or Qwen, where replaying synthetic data could keep the model
from forgetting what it has learned.

\subsection*{AI use statement}
In this work, the method itself uses a generative AI model: Claude Sonnet~4 writes and rewrites the data
generating programs during search (Section~\ref{sec:method}). We also used generative AI tools to implement
experiment code, run and analyse experiments, and assist in writing and editing the paper. All AI-assisted
code, analyses and text were reviewed by the authors; every number in the paper is computed by a script from
the raw run logs. We take responsibility for the final content of this work, including text, claims and
artifacts produced with the aid of generative AI.

\subsection*{Ethics statement}
This work trains small models on public datasets (TinyStories, WikiText-103, CodeParrot and CodeSearchNet) and on
procedurally generated puzzles; it involves no human subjects or personal data, and the synthetic pretraining data
consists only of abstract symbols (Appendix~\ref{app:contract}).

\subsection*{Reproducibility statement}
The model, optimiser, search schedule and evaluation protocol are specified in Section~\ref{sec:setup} and
Appendix~\ref{app:setup}; the generator contract and prompt in Appendices~\ref{app:contract}
and~\ref{app:prompt}; dataset sources and splits in Appendix~\ref{app:data}. Code, all optimized programs and the
script that computes every table from the run logs will be released.

\bibliography{references}

\begin{thebibliography}{35}
\providecommand{\natexlab}[1]{#1}
\providecommand{\url}[1]{\texttt{#1}}
\expandafter\ifx\csname urlstyle\endcsname\relax
  \providecommand{\doi}[1]{doi: #1}\else
  \providecommand{\doi}{doi: \begingroup \urlstyle{rm}\Url}\fi

\bibitem[{Anthropic}(2025)]{anthropic2025claude4}
{Anthropic}.
\newblock System card: {C}laude {O}pus 4 \& {C}laude {S}onnet 4.
\newblock \url{https://www.anthropic.com/claude-4-system-card}, 2025.

\bibitem[Barab{\'a}si et~al.(2023)Barab{\'a}si, Beynon, Katona, and
  Perez-Nieves]{barabasi2023complex}
D{\'a}niel~L. Barab{\'a}si, Taliesin Beynon, {\'A}d{\'a}m Katona, and Nicolas
  Perez-Nieves.
\newblock Complex computation from developmental priors.
\newblock \emph{Nature Communications}, 14\penalty0 (1):\penalty0 2226, 2023.

\bibitem[{CodeParrot}(2022)]{codeparrot}
{CodeParrot}.
\newblock {CodeParrot} clean python dataset.
\newblock
  \url{https://huggingface.co/datasets/codeparrot/codeparrot-clean-train},
  2022.

\bibitem[Cowsik et~al.(2026)Cowsik, Dolev, Li, De~Luca, Cohen, Goodman, and
  Levine]{cowsik2026selfplay}
Aditya Cowsik, Kfir Dolev, Michael~Y. Li, G.~Bruno De~Luca, Nourya Cohen,
  Noah~D. Goodman, and Yoav Levine.
\newblock Self-play pretraining with zero data.
\newblock \emph{arXiv preprint arXiv:2609.30063}, 2026.

\bibitem[Eldan \& Li(2023)Eldan and Li]{eldan2023tinystories}
Ronen Eldan and Yuanzhi Li.
\newblock {TinyStories}: How small can language models be and still speak
  coherent {E}nglish?
\newblock \emph{arXiv preprint arXiv:2305.07759}, 2023.

\bibitem[Elistratov et~al.(2026)Elistratov, Barannikov, Ivanov, Khrulkov,
  Konushin, Kuznetsov, and Zhemchuzhnikov]{elistratov2026cadevolve}
Maksim Elistratov, Marina Barannikov, Gregory Ivanov, Valentin Khrulkov, Anton
  Konushin, Andrey Kuznetsov, and Dmitrii Zhemchuzhnikov.
\newblock {CADEvolve}: Creating realistic {CAD} via program evolution.
\newblock \emph{arXiv preprint arXiv:2602.16317}, 2026.

\bibitem[Finn et~al.(2017)Finn, Abbeel, and Levine]{finn2017maml}
Chelsea Finn, Pieter Abbeel, and Sergey Levine.
\newblock Model-agnostic meta-learning for fast adaptation of deep networks.
\newblock In \emph{International Conference on Machine Learning (ICML)}, 2017.

\bibitem[Finzi et~al.(2026)Finzi, Qiu, Jiang, Izmailov, Kolter, and
  Wilson]{finzi2026epiplexity}
Marc Finzi, Shikai Qiu, Yiding Jiang, Pavel Izmailov, J.~Zico Kolter, and
  Andrew~Gordon Wilson.
\newblock From entropy to epiplexity: Rethinking information for
  computationally bounded intelligence.
\newblock \emph{arXiv preprint arXiv:2601.03220}, 2026.

\bibitem[Gandhi et~al.(2024)Gandhi, Lee, Grand, Liu, Cheng, Sharma, and
  Goodman]{gandhi2024sos}
Kanishk Gandhi, Denise Lee, Gabriel Grand, Muxin Liu, Winson Cheng, Archit
  Sharma, and Noah~D. Goodman.
\newblock Stream of search ({SoS}): Learning to search in language.
\newblock \emph{arXiv preprint arXiv:2404.03683}, 2024.

\bibitem[Grau-Moya et~al.(2024)Grau-Moya, Genewein, Hutter, Orseau,
  Del{\'e}tang, Catt, Ruoss, Wenliang, Mattern, Aitchison, and
  Veness]{graumoya2024universal}
Jordi Grau-Moya, Tim Genewein, Marcus Hutter, Laurent Orseau, Gr{\'e}goire
  Del{\'e}tang, Elliot Catt, Anian Ruoss, Li~Kevin Wenliang, Christopher
  Mattern, Matthew Aitchison, and Joel Veness.
\newblock Learning universal predictors.
\newblock In \emph{International Conference on Machine Learning (ICML)}, 2024.

\bibitem[He et~al.(2019)He, Girshick, and Doll{\'a}r]{he2019rethinking}
Kaiming He, Ross Girshick, and Piotr Doll{\'a}r.
\newblock Rethinking {ImageNet} pre-training.
\newblock In \emph{IEEE/CVF International Conference on Computer Vision
  (ICCV)}, 2019.

\bibitem[Hernandez et~al.(2021)Hernandez, Kaplan, Henighan, and
  McCandlish]{hernandez2021transfer}
Danny Hernandez, Jared Kaplan, Tom Henighan, and Sam McCandlish.
\newblock Scaling laws for transfer.
\newblock \emph{arXiv preprint arXiv:2102.01293}, 2021.

\bibitem[Ho et~al.(2019)Ho, Liang, Stoica, Abbeel, and Chen]{ho2019pba}
Daniel Ho, Eric Liang, Ion Stoica, Pieter Abbeel, and Xi~Chen.
\newblock Population based augmentation: Efficient learning of augmentation
  policy schedules.
\newblock In \emph{International Conference on Machine Learning (ICML)}, 2019.

\bibitem[Hu et~al.(2025)Hu, Petty, Shi, Merrill, and Linzen]{hu2025circuits}
Michael~Y. Hu, Jackson Petty, Chuan Shi, William Merrill, and Tal Linzen.
\newblock Between circuits and {C}homsky: Pre-pretraining on formal languages
  imparts linguistic biases.
\newblock In \emph{Annual Meeting of the Association for Computational
  Linguistics (ACL)}, 2025.

\bibitem[Husain et~al.(2019)Husain, Wu, Gazit, Allamanis, and
  Brockschmidt]{husain2019codesearchnet}
Hamel Husain, Ho-Hsiang Wu, Tiferet Gazit, Miltiadis Allamanis, and Marc
  Brockschmidt.
\newblock {CodeSearchNet} challenge: Evaluating the state of semantic code
  search.
\newblock \emph{arXiv preprint arXiv:1909.09436}, 2019.

\bibitem[Jiang et~al.(2026)Jiang, Shinnick, van~den Hengel, Saratchandran, and
  Teney]{jiang2026procedural}
Liangze Jiang, Zachary Shinnick, Anton van~den Hengel, Hemanth Saratchandran,
  and Damien Teney.
\newblock Procedural pretraining: Warming up language models with abstract
  data.
\newblock \emph{arXiv preprint arXiv:2601.21725}, 2026.

\bibitem[Lake \& Baroni(2018)Lake and Baroni]{lake2018scan}
Brenden~M. Lake and Marco Baroni.
\newblock Generalization without systematicity: On the compositional skills of
  sequence-to-sequence recurrent networks.
\newblock In \emph{International Conference on Machine Learning (ICML)}, 2018.

\bibitem[Lee et~al.(2026)Lee, Han, Kumar, and Agrawal]{lee2026nca}
Dan Lee, Seungwook Han, Akarsh Kumar, and Pulkit Agrawal.
\newblock Training language models via neural cellular automata.
\newblock \emph{arXiv preprint arXiv:2603.10055}, 2026.

\bibitem[Li et~al.(2025)Li, Yu, and Xiong]{li2025montessori}
Xiaochuan Li, Zichun Yu, and Chenyan Xiong.
\newblock Montessori-instruct: Generate influential training data tailored for
  student learning.
\newblock In \emph{International Conference on Learning Representations
  (ICLR)}, 2025.

\bibitem[Li et~al.(2022)Li, Choi, Chung, Kushman, Schrittwieser, Leblond,
  Eccles, Keeling, Gimeno, Dal~Lago, Hubert, Choy, de~Masson~d'Autume,
  Babuschkin, Chen, Huang, Welbl, Gowal, Cherepanov, Molloy, Mankowitz,
  Sutherland~Robson, Kohli, de~Freitas, Kavukcuoglu, and
  Vinyals]{li2022alphacode}
Yujia Li, David Choi, Junyoung Chung, Nate Kushman, Julian Schrittwieser,
  R{\'e}mi Leblond, Tom Eccles, James Keeling, Felix Gimeno, Agustin Dal~Lago,
  Thomas Hubert, Peter Choy, Cyprien de~Masson~d'Autume, Igor Babuschkin,
  Xinyun Chen, Po-Sen Huang, Johannes Welbl, Sven Gowal, Alexey Cherepanov,
  James Molloy, Daniel~J. Mankowitz, Esme Sutherland~Robson, Pushmeet Kohli,
  Nando de~Freitas, Koray Kavukcuoglu, and Oriol Vinyals.
\newblock Competition-level code generation with {AlphaCode}.
\newblock \emph{Science}, 378\penalty0 (6624):\penalty0 1092--1097, 2022.

\bibitem[Liang et~al.(2024)Liang, Wang, Wang, Bastani, Jayaraman, and
  Ma]{liang2024eurekaverse}
William Liang, Sam Wang, Hung-Ju Wang, Osbert Bastani, Dinesh Jayaraman, and
  Yecheng~Jason Ma.
\newblock Eurekaverse: Environment curriculum generation via large language
  models.
\newblock In \emph{Conference on Robot Learning (CoRL)}, 2024.

\bibitem[Merity et~al.(2017)Merity, Xiong, Bradbury, and
  Socher]{merity2017pointer}
Stephen Merity, Caiming Xiong, James Bradbury, and Richard Socher.
\newblock Pointer sentinel mixture models.
\newblock In \emph{International Conference on Learning Representations
  (ICLR)}, 2017.

\bibitem[Mi et~al.(2026)Mi, Shan, Huang, Qin, Xie, Qiao, Liu, Zhou, and
  Liu]{mi2026dataevolve}
Tiantian Mi, Dongming Shan, Zhen Huang, Yiwei Qin, Muhang Xie, Yuxuan Qiao,
  Yixiu Liu, Chenyang Zhou, and Pengfei Liu.
\newblock Data {D}arwinism part {II}: {DataEvolve} -- {AI} can autonomously
  evolve pretraining data curation.
\newblock \emph{arXiv preprint arXiv:2603.14420}, 2026.

\bibitem[Novikov et~al.(2025)Novikov, V{\~u}, Eisenberger, Dupont, Huang,
  Wagner, Shirobokov, Kozlovskii, Ruiz, Mehrabian, Kumar, See, Chaudhuri,
  Holland, Davies, Nowozin, Kohli, and Balog]{novikov2025alphaevolve}
Alexander Novikov, Ng{\^a}n V{\~u}, Marvin Eisenberger, Emilien Dupont, Po-Sen
  Huang, Adam~Zsolt Wagner, Sergey Shirobokov, Borislav Kozlovskii, Francisco
  J.~R. Ruiz, Abbas Mehrabian, M.~Pawan Kumar, Abigail See, Swarat Chaudhuri,
  George Holland, Alex Davies, Sebastian Nowozin, Pushmeet Kohli, and Matej
  Balog.
\newblock {AlphaEvolve}: A coding agent for scientific and algorithmic
  discovery.
\newblock \emph{arXiv preprint arXiv:2506.13131}, 2025.

\bibitem[Papadimitriou \& Jurafsky(2020)Papadimitriou and
  Jurafsky]{papadimitriou2020music}
Isabel Papadimitriou and Dan Jurafsky.
\newblock Learning music helps you read: Using transfer to study linguistic
  structure in language models.
\newblock In \emph{Conference on Empirical Methods in Natural Language
  Processing (EMNLP)}, 2020.

\bibitem[Romera-Paredes et~al.(2024)Romera-Paredes, Barekatain, Novikov, Balog,
  Kumar, Dupont, Ruiz, Ellenberg, Wang, Fawzi, Kohli, and
  Fawzi]{romeraparedes2024funsearch}
Bernardino Romera-Paredes, Mohammadamin Barekatain, Alexander Novikov, Matej
  Balog, M.~Pawan Kumar, Emilien Dupont, Francisco J.~R. Ruiz, Jordan~S.
  Ellenberg, Pengming Wang, Omar Fawzi, Pushmeet Kohli, and Alhussein Fawzi.
\newblock Mathematical discoveries from program search with large language
  models.
\newblock \emph{Nature}, 625:\penalty0 468--475, 2024.

\bibitem[Schkufza et~al.(2013)Schkufza, Sharma, and Aiken]{schkufza2013stoke}
Eric Schkufza, Rahul Sharma, and Alex Aiken.
\newblock Stochastic superoptimization.
\newblock In \emph{International Conference on Architectural Support for
  Programming Languages and Operating Systems (ASPLOS)}, 2013.

\bibitem[Shinnick et~al.(2025)Shinnick, Jiang, Saratchandran, Teney, and
  van~den Hengel]{shinnick2025see}
Zachary Shinnick, Liangze Jiang, Hemanth Saratchandran, Damien Teney, and Anton
  van~den Hengel.
\newblock Can you learn to see without images? {P}rocedural warm-up for vision
  transformers.
\newblock \emph{arXiv preprint arXiv:2511.13945}, 2025.

\bibitem[Stanley et~al.(2009)Stanley, D'Ambrosio, and
  Gauci]{stanley2009hypercube}
Kenneth~O. Stanley, David~B. D'Ambrosio, and Jason Gauci.
\newblock A hypercube-based encoding for evolving large-scale neural networks.
\newblock \emph{Artificial Life}, 15\penalty0 (2):\penalty0 185--212, 2009.

\bibitem[Such et~al.(2020)Such, Rawal, Lehman, Stanley, and Clune]{such2020gtn}
Felipe~Petroski Such, Aditya Rawal, Joel Lehman, Kenneth~O. Stanley, and Jeff
  Clune.
\newblock Generative teaching networks: Accelerating neural architecture search
  by learning to generate synthetic training data.
\newblock In \emph{International Conference on Machine Learning (ICML)}, 2020.

\bibitem[Trinh et~al.(2024)Trinh, Wu, Le, He, and
  Luong]{trinh2024alphageometry}
Trieu~H. Trinh, Yuhuai Wu, Quoc~V. Le, He~He, and Thang Luong.
\newblock Solving olympiad geometry without human demonstrations.
\newblock \emph{Nature}, 625:\penalty0 476--482, 2024.

\bibitem[Valmeekam et~al.(2023)Valmeekam, Marquez, Olmo, Sreedharan, and
  Kambhampati]{valmeekam2023planbench}
Karthik Valmeekam, Matthew Marquez, Alberto Olmo, Sarath Sreedharan, and
  Subbarao Kambhampati.
\newblock {PlanBench}: An extensible benchmark for evaluating large language
  models on planning and reasoning about change.
\newblock In \emph{Advances in Neural Information Processing Systems (NeurIPS),
  Datasets and Benchmarks Track}, 2023.

\bibitem[Wu et~al.(2021)Wu, Rabe, Li, Ba, Grosse, and Szegedy]{wu2021lime}
Yuhuai Wu, Markus~N. Rabe, Wenda Li, Jimmy Ba, Roger~B. Grosse, and Christian
  Szegedy.
\newblock {LIME}: Learning inductive bias for primitives of mathematical
  reasoning.
\newblock In \emph{International Conference on Machine Learning (ICML)}, 2021.

\bibitem[Zhang et~al.(2025)Zhang, Patel, Rizvi, Liu, He, Karbasi, Zappala, and
  van Dijk]{zhang2025edge}
Shiyang Zhang, Aakash Patel, Syed~A. Rizvi, Nianchen Liu, Sizhuang He, Amin
  Karbasi, Emanuele Zappala, and David van Dijk.
\newblock Intelligence at the edge of chaos.
\newblock In \emph{International Conference on Learning Representations
  (ICLR)}, 2025.

\bibitem[Zhao et~al.(2025)Zhao, Wu, Yue, Wu, Xu, Yue, Lin, Wang, Wu, Zheng, and
  Huang]{zhao2025absolute}
Andrew Zhao, Yiran Wu, Yang Yue, Tong Wu, Quentin Xu, Yang Yue, Matthieu Lin,
  Shenzhi Wang, Qingyun Wu, Zilong Zheng, and Gao Huang.
\newblock Absolute zero: Reinforced self-play reasoning with zero data.
\newblock \emph{arXiv preprint arXiv:2505.03335}, 2025.

\end{thebibliography}
\bibliographystyle{iclr2027_conference}

\appendix
\raggedbottom
\section{Algorithm}\label{app:alg}
\begin{figure}[H]
  \centering
  \resizebox{\textwidth}{!}{\definecolor{progred}{HTML}{C0392B}\definecolor{progfill}{HTML}{FBEAEA}
\definecolor{datafill}{HTML}{DCE6F2}\definecolor{dataline}{HTML}{5B7BA6}
\definecolor{netfill}{HTML}{FDF1DC}\definecolor{netline}{HTML}{D9A44A}
\definecolor{evalfill}{HTML}{D5EEEE}\definecolor{evalline}{HTML}{3A9A9A}
\definecolor{llmfill}{HTML}{ECE6F5}\definecolor{llmline}{HTML}{7A55B0}
\definecolor{inkgray}{HTML}{52514E}
\begin{tikzpicture}[
  font=\sffamily\footnotesize, >={Stealth[length=4pt]}, node distance=4mm,
  box/.style={draw, rounded corners=2pt, line width=0.6pt, align=center, minimum height=9mm, inner sep=3pt},
  net/.style={box, fill=netfill, draw=netline, minimum width=19mm},
  ev/.style={box, fill=evalfill, draw=evalline, minimum width=17mm},
  llm/.style={box, fill=llmfill, draw=llmline, minimum width=21mm},
  cyl/.style={cylinder, shape border rotate=90, aspect=0.22, draw=dataline, fill=datafill, line width=0.6pt,
              align=center, minimum width=13mm, minimum height=9mm, inner sep=1pt},
  flow/.style={->, line width=0.6pt, draw=black!75},
  back/.style={->, line width=0.6pt, draw=progred},
  note/.style={font=\sffamily\scriptsize\itshape, text=inkgray}]
  \node[draw=progred, fill=progfill, line width=0.7pt, rounded corners=2pt, minimum width=33mm, text width=33mm, minimum height=15mm,
        align=left, inner sep=3pt, text depth=-3.5mm] (prog) {\texttt{\scriptsize def generate\_stream():}\\[-1pt]\texttt{\scriptsize\ \ \ \ ...}};
  \node[fill=progred, text=white, font=\ttfamily\scriptsize\bfseries, anchor=north west, inner sep=2.2pt,
        minimum width=33mm, text width=31mm, align=left, rounded corners=2pt] at (prog.north west) {datagen.py ($\pi$)};
  \node[cyl, right=5mm of prog] (syn) {Synthetic\\data};
  \node[net, right=of syn] (pre) {Transformer\\$\theta(\pi)$};
  \node[net, right=of pre] (ft) {Transformer\\$\theta(\pi,\mathcal{D}_{\text{train}})$};
  \node[ev, right=of ft] (eval) {Evaluate\\$\hat{J}(\pi)$};
  \node[cyl, above=3mm of ft] (dtr) {$\mathcal{D}_{\text{train}}$};
  \node[cyl, above=3mm of eval] (dva) {$\mathcal{D}_{\text{val}}$};
  \node[ev, below=7mm of eval, minimum width=20mm] (acc) {Accept if\\$\hat{J}\le\hat{J}^{*}$};
  \node[llm, left=12mm of acc] (hist) {Programs and\\scores so far};
  \node[llm] (lm) at (prog.south |- hist) {LLM\\{\scriptsize(program rewriter)}};
  \draw[flow] (prog) -- (syn); \draw[flow] (syn) -- (pre); \draw[flow] (pre) -- (ft); \draw[flow] (ft) -- (eval);
  \draw[flow] (dtr) -- (ft); \draw[flow] (dva) -- (eval); \draw[flow] (eval) -- (acc);
  \draw[flow] (acc) -- (hist); \draw[flow] (hist) -- (lm);
  \draw[back] (lm) -- node[note, right, pos=0.5] {rewrite} (prog);
  \draw[->, line width=0.6pt, draw=netline, dashed] (acc.south) |- ++(0,-4.5mm) -| node[note, pos=0.23, below]
       {accept: next candidate continues pretraining $\theta^{*}$ \;\; reject: roll back to $\theta^{*}, \pi^{*}$} (pre.south);
\end{tikzpicture}}
  \caption{\method{}. The LLM rewrites the generator program; the rewrite is scored by continuing to pretrain
  the one network kept for the whole search, fine-tuning a copy on $\mathcal{D}_\text{train}$ and evaluating it
  on $\mathcal{D}_\text{val}$. An improvement is accepted; otherwise both the program and the network are rolled
  back to the last accepted state.}
  \label{fig:method}
\end{figure}
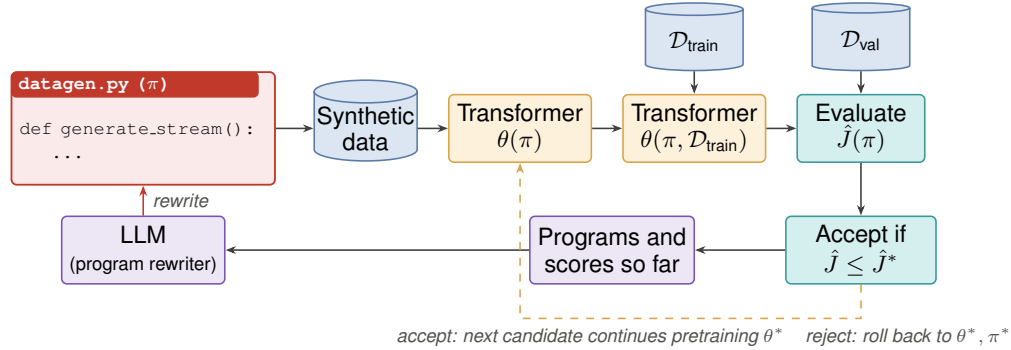
\begin{algorithm}[H]
\caption{\method{}}\label{alg:precocial}
\small
\begin{algorithmic}[1]
\Require Downstream data $\mathcal{D}$, LLM, rounds $K$, steps $T_0, T_c$
\State $\pi \gets \text{LLM}(\text{contract}, \mathcal{D}_\text{sample})$ \Comment{first iteration}
\State $\theta \gets \text{Pretrain}(\text{RandomInit}(), \text{Execute}(\pi), T_0)$
\State $\hat{J}^* \gets \text{Eval}(\text{Finetune}(\theta, \mathcal{D}_\text{train}), \mathcal{D}_\text{val})$;\quad $(\pi^*, \theta^*) \gets (\pi, \theta)$
\For{$k = 1, \dots, K-1$}
  \State $\pi_k \gets \text{LLM}(\pi^*, \hat{J}^*, \mathcal{D}_\text{sample})$ \Comment{rewrite the accepted program}
  \State $\theta \gets \text{ContinuePretrain}(\theta^*, \text{Execute}(\pi_k), T_c)$ \Comment{resume the one network}
  \State $\hat{J}_k \gets \text{Eval}(\text{Finetune}(\theta, \mathcal{D}_\text{train}), \mathcal{D}_\text{val})$ \Comment{with early stopping}
  \If{$\hat{J}_k \le \hat{J}^*$} $(\pi^*, \theta^*, \hat{J}^*) \gets (\pi_k, \theta, \hat{J}_k)$ \Comment{accept}
  \EndIf \Comment{otherwise roll back to $(\pi^*, \theta^*)$}
\EndFor
\State \Return $\pi^*, \theta^*$
\end{algorithmic}
\end{algorithm}

\section{Further experimental setup details}\label{app:setup}
\textbf{Model.} Vocabulary 260 (256 bytes + PAD/BOS/EOS/SEP), 3 layers, 4 attention heads, 128-dimensional
embedding, 512-position context (1{,}024 for Blocksworld), $\approx$0.69M parameters. All weights are randomly
initialised; embeddings are not reinitialised between pretraining and fine-tuning.

\textbf{Pretraining.} 2{,}000 streams per round, 50 sentences per stream, truncated to the context length. Round
0: 15{,}000 steps from scratch; later rounds: 5{,}000 continued steps from the last accepted checkpoint. A
ten-round search therefore performs 60{,}000 pretraining steps, and its kept network carries between 15{,}000 and
60{,}000 of them, depending on how many rounds were accepted. Batch size 8, AdamW, learning rate
$3\times10^{-4}$, gradient clipping at 1.0.

\begin{figure}[H]
  \begin{minipage}[c]{0.5\textwidth}
    \centering
    \includegraphics[width=\textwidth]{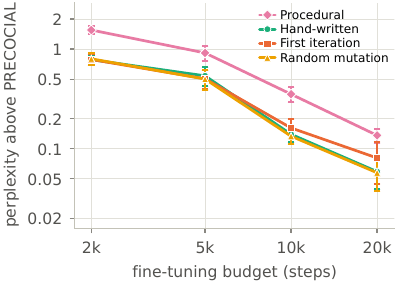}
  \end{minipage}\hfill
  \begin{minipage}[c]{0.47\textwidth}
    \caption{Perplexity above \method{} at each fine-tuning budget, pooled over the four datasets under matched
    pretraining (best validation perplexity within the budget; mean\,$\pm$\,s.e.m.\ over 36 paired runs;
    logarithmic axes). \method{} is ahead of every other curriculum at every budget shown; the lead is largest
    at small budgets and narrows as fine-tuning proceeds. See also Table~\ref{tab:budget-matched}.}
    \label{fig:budget}
  \end{minipage}
\end{figure}

\textbf{Fine-tuning.} Up to 10{,}000 steps, learning rate $3\times10^{-4}$, batch size 8, early stopping with
patience 5 evaluations every 500 steps; the checkpoint with the lowest validation perplexity is tested. A second,
longer fine-tune of the kept pretrained network (early stopping with patience 10; every run continued well past 20k
steps) supplies the validation curve from which we read the best perplexity reached within each fine-tuning budget
from 2k to 20k steps (Figures~\ref{fig:budget} and~\ref{fig:headstart}, Appendix~\ref{app:tables}). For the
puzzles the loss and the perplexity are computed on the answer tokens only.

\textbf{Choice of budgets.} A search evaluates ten candidates, each with its own pretraining and fine-tuning,
so each evaluation must be short. We fine-tune for 10k steps (about eight passes over a 10k-document training set
at batch size 8), which keeps a ten-candidate search under one GPU-hour on the 0.69M model
(Appendix~\ref{app:cost}). It is a budget, not convergence: validation perplexity is still falling at 10k steps
in 96\% of the fine-tunes run during our searches. Because the search's fitness is measured at this budget,
results at 10k steps evaluate exactly what the search optimizes; Figure~\ref{fig:budget} and
Table~\ref{tab:budget-matched} report 2k to 20k steps as well. Ten rounds keep the LLM cost near \$1 per search,
and the program kept after round 5 already captures most of the gain (Table~\ref{tab:lang}), so more rounds would
add little. Both values were set once, before the experiments reported here, and used unchanged for every
dataset, task, baseline and model size.

\textbf{Search.} Ten rounds; a candidate is accepted if its validation perplexity is no worse than the best so
far (tolerance 0), and otherwise both the program and the pretrained checkpoint are rolled back. The LLM (Claude
Sonnet~4 via the OpenRouter API) samples the first-iteration program at temperature 0.9 and rewrites at temperature 0.7.
FunSearch uses 4 islands and prompts with the best programs of one island; AlphaEvolve uses one database,
fitness-proportional parent sampling and 2 inspiration programs per prompt. As published, both pretrain a fresh
network from random initialisation for every candidate (15{,}000 steps, then the fine-tuning above). All three
methods make one LLM call per evaluation, plus a retry whenever a returned program fails validation.

\textbf{Matched pretraining.} For every comparison of curricula (Tables~\ref{tab:lang}, \ref{tab:planning}
and~\ref{tab:transfer}), the kept program of \method{} and of every baseline pretrains a fresh network for exactly
60{,}000 steps on 2{,}000 streams, followed by the fine-tuning above. For language, each of the three kept programs
per method (one per search seed) is trained with three training seeds, the search seed and two others, giving 9 runs per cell
of Tables~\ref{tab:lang} and~\ref{tab:transfer}; no pretraining uses the same nine training seeds, and runs are
paired by training seed. Blocksworld is fine-tuned for the full 10{,}000 steps without
early stopping: its validation perplexity plateaus early, and patience-5 stopping ends fine-tuning at 1{,}000 to
3{,}500 steps, above the perplexity at which plans start to succeed (Appendix~\ref{app:planning}).

\section{Generator program specification}\label{app:contract}
The generator contract requires programs to define:
\begin{verbatim}
def generate_stream(num_sentences: int) -> List[Tuple[str, str]]:
    """Returns list of (rule_name, content_string) pairs.
    content_string must use only abstract ASCII symbols."""
\end{verbatim}
All content strings are drawn from the following vocabulary: uppercase letters (A--Z), digits (0--9), brackets
(\texttt{()[]<>\{\}}), operators (\texttt{+-*/=\^{}\%\~{}}), separators (\texttt{|;,.!@\#\$ }), and arrows
(\texttt{->}, \texttt{=>}). English words and multi-character token names are forbidden by a validation step that
checks for common English words before accepting any generated program. The rule name returned with each
content string tags the stream and is used for the rule analysis (Section~\ref{sec:exp-rules}).

\section{Prompt structure}\label{app:prompt}
\textbf{First iteration.} The LLM is asked to design synthetic pretraining data for a small GPT-2 model whose
transformer layers (not its token embeddings' meaning) will transfer to the downstream task, and receives the
generator contract, which lists the rule families of Table~\ref{tab:init} with their literature justifications,
the task type, and 500 bytes of downstream training data.

\textbf{Each later round.} The prompt contains, in order: (1) the round number and a note that the network is
continued from the previous round rather than retrained; (2) the current validation perplexity and the best so
far; (3) the validation perplexity along the fine-tuning curves of the current and previous candidates;
(4) 300 bytes of downstream training data; (5) the current program in a Python code fence; (6) the generator
contract; and (7) the instruction: add any number of new rule types, remove any rules that are unhelpful, and
modify existing generators and their weights, with no cap on the size of the change (the prompt notes, as
information only, that large distribution shifts can cause forgetting in a continually trained network). The LLM
returns the complete program. For the task-hidden ablation (Section~\ref{sec:exp-rules}) the downstream sample is
replaced by a notice that it is withheld.

\section{Datasets}\label{app:data}
\begin{table}[H]
  \caption{Dataset splits. Language and code datasets are byte-encoded as raw text with no filtering at training
  time beyond truncation to the context. Puzzles are generated with fixed seeds, with a reference answer that is
  optimal for the move puzzles; any answer the checker accepts counts as solved. $^\dagger$The first 200 of 1{,}000
  test problems, decoded greedily.}
  \label{tab:data}
  \centering\small
  \begin{adjustbox}{max width=\linewidth}\begin{tabular}{@{}lrrrl@{}}
    \toprule
    Dataset & Train & Val & Test & Source / checker \\
    \midrule
    TinyStories & 7{,}818 & 868 & 155 & HF train / validation, $\le$512 characters \\
    WikiText-103 & 10{,}000 & 1{,}111 & 1{,}241 & HF train / HF test (raw) \\
    CodeParrot & 10{,}000 & 2{,}499 & 2{,}499 & codeparrot-clean-train, held-out tail \\
    CodeSearchNet & 10{,}000 & 1{,}000 & 1{,}000 & Python functions, held out \\
    \midrule
    Blocksworld (9--12 blocks) & 10{,}000 & 1{,}000 & 200 & PDDL; plan replayed in a simulator$^\dagger$ \\
    8-puzzle & 5{,}000 & 200 & 500 & 10--20 moves from the goal; reach it \\
    Sokoban ($7{\times}7$, 2 boxes) & 5{,}000 & 200 & 500 & push both boxes onto targets \\
    Countdown (4 numbers) & 5{,}000 & 200 & 500 & each number once; equals the target \\
    \bottomrule
  \end{tabular}\end{adjustbox}
\end{table}

\section{Additional results}\label{app:tables}
\subsection{Every fine-tuning budget}

\begin{table}[H]
  \caption{Matched pretraining: perplexity above \method{} within each fine-tuning budget, pooled over the four
  datasets, 36 paired runs; $p$ in parentheses; bold: $p<.05$. Each entry is the best validation perplexity
  reached within the budget along a single long fine-tuning run. These are the values plotted in
  Figure~\ref{fig:budget}; they differ slightly from the 10k-step column of Table~\ref{tab:lang}, which is the
  test perplexity of a separate 10k-step fine-tune.}
  \label{tab:budget-matched}
  \centering\small\setlength{\tabcolsep}{4pt}
  \begin{adjustbox}{max width=\linewidth}\begin{tabular}{@{}lcccc@{}}
    \toprule
    Baseline & 2k & 5k & 10k & 20k \\
    \midrule
    First iteration & \textbf{$+$0.786} {\scriptsize($<$.001)} & \textbf{$+$0.504} {\scriptsize($<$.001)} & \textbf{$+$0.162} {\scriptsize($<$.001)} & \textbf{$+$0.081} {\scriptsize(.032)} \\
Random mutation & \textbf{$+$0.800} {\scriptsize($<$.001)} & \textbf{$+$0.501} {\scriptsize($<$.001)} & \textbf{$+$0.134} {\scriptsize($<$.001)} & \textbf{$+$0.058} {\scriptsize(.006)} \\
Procedural & \textbf{$+$1.554} {\scriptsize($<$.001)} & \textbf{$+$0.915} {\scriptsize($<$.001)} & \textbf{$+$0.353} {\scriptsize($<$.001)} & \textbf{$+$0.137} {\scriptsize($<$.001)} \\
Hand-written ($\pi_0$) & \textbf{$+$0.784} {\scriptsize($<$.001)} & \textbf{$+$0.540} {\scriptsize($<$.001)} & \textbf{$+$0.140} {\scriptsize($<$.001)} & \textbf{$+$0.059} {\scriptsize(.005)} \\
No pretraining & \textbf{$+$1.685} {\scriptsize($<$.001)} & \textbf{$+$0.900} {\scriptsize($<$.001)} & \textbf{$+$0.476} {\scriptsize($<$.001)} & \textbf{$+$0.307} {\scriptsize(.001)} \\
 
    \bottomrule
  \end{tabular}\end{adjustbox}
\end{table}

\begin{table}[H]
  \caption{Fine-tuning steps saved, matched pretraining. For each budget $k$, the share of a baseline's fine-tuning
  steps that \method{} needs to reach the same validation perplexity: $k$ divided by the steps the baseline needs to
  reach the best perplexity \method{} attains within $k$ steps, median over the 36 runs (4 datasets $\times$ 9). Lower means a larger saving. Curves are evaluated every 500 steps.}
  \label{tab:savings}
  \centering\small\setlength{\tabcolsep}{6pt}
  \begin{adjustbox}{max width=\linewidth}\begin{tabular}{@{}lcccc@{}}
    \toprule
    Baseline & 2k & 5k & 10k & 20k \\
    \midrule
    First iteration & 80\% & 83\% & 91\% & 93\% \\
Random mutation & 80\% & 91\% & 91\% & 94\% \\
Procedural & 67\% & 74\% & 80\% & 82\% \\
Hand-written ($\pi_0$) & 80\% & 83\% & 89\% & 91\% \\
No pretraining & 57\% & 77\% & 83\% & 83\% \\
 
    \bottomrule
  \end{tabular}\end{adjustbox}
\end{table}

\subsection{The ten-round loop without matched pretraining}\label{app:loop}
The tables below evaluate each method with the network its own ten-round loop produced, instead of retraining its
kept program on a fresh network. Methods then differ in accumulated pretraining: the kept checkpoints carry
42{,}500 pretraining steps on average for \method{} against 37{,}100--38{,}800 for the baselines
\unskip. \method{} is lowest on every dataset, and every pooled difference is
significant.
\begin{table}[H]
  \caption{End of the ten-round loop: test perplexity after a 10k-step fine-tune, mean\,$\pm$\,SD over 3 seeds;
  last column pooled over the four datasets (positive: \method{} lower).}
  \label{tab:lang-loop}
  \centering\small\setlength{\tabcolsep}{3.3pt}
  \begin{adjustbox}{max width=\linewidth}\begin{tabular}{@{}lccccc@{}}
    \toprule
    Pretraining & WikiText-103 & TinyStories & CodeParrot & CodeSearchNet & $\Delta$ vs \method{} \\
    \midrule
    \rowcolor{hl}\method{} & 5.312\std{.146} & 2.825\std{.040} & 3.176\std{.086} & 4.035\std{.108} & -- \\
First iteration & 5.493\std{.045} & 2.885\std{.013} & 3.291\std{.025} & 4.278\std{.114} & \textbf{$+$0.150} {\scriptsize($<$.001)} \\
Random mutation & 5.479\std{.027} & 2.904\std{.022} & 3.332\std{.051} & 4.418\std{.414} & \textbf{$+$0.196} {\scriptsize(.005)} \\
Procedural & 5.813\std{.080} & 2.964\std{.032} & 3.493\std{.130} & 4.509\std{.077} & \textbf{$+$0.358} {\scriptsize($<$.001)} \\
Hand-written ($\pi_0$) & 5.508\std{.082} & 2.891\std{.021} & 3.311\std{.015} & 4.288\std{.084} & \textbf{$+$0.162} {\scriptsize($<$.001)} \\
No pretraining & 5.585\std{.055} & 2.939\std{.077} & 3.205\std{.004} & 5.676\std{.149} & \textbf{$+$0.514} {\scriptsize(.026)} \\
 
    \bottomrule
  \end{tabular}\end{adjustbox}
\end{table}
\begin{table}[H]
  \caption{Transfer with the networks produced by the WikiText-103 loops (no matched pretraining): test perplexity
  after a 10k-step fine-tune on the target, mean\,$\pm$\,SD over 3 seeds; last column pooled over the three targets.}
  \label{tab:transfer-loop}
  \centering\small\setlength{\tabcolsep}{3pt}
  \begin{adjustbox}{max width=\linewidth}\begin{tabular}{@{}lcccc@{}}
    \toprule
    WikiText-103 program & TinyStories & CodeParrot & CodeSearchNet & $\Delta$ vs \method{} \\
    \midrule
    \rowcolor{hl}\method{} & 2.840\std{.067} & 3.206\std{.082} & 4.161\std{.218} & -- \\
First iteration & 2.917\std{.020} & 3.350\std{.060} & 5.565\std{.065} & \textbf{$+$0.542} {\scriptsize(.041)} \\
Random mutation & 2.915\std{.048} & 3.344\std{.039} & 5.093\std{.723} & $+$0.382 {\scriptsize(.072)} \\
Procedural & 3.010\std{.037} & 3.453\std{.015} & 4.591\std{.194} & \textbf{$+$0.283} {\scriptsize(.006)} \\
Hand-written ($\pi_0$) & 2.922\std{.056} & 3.357\std{.054} & 4.452\std{.171} & \textbf{$+$0.175} {\scriptsize(.002)} \\
No pretraining & 2.908\std{.047} & 3.200\std{.016} & 5.704\std{.099} & $+$0.535 {\scriptsize(.070)} \\
\midrule
\textit{Optimized on the target} & 2.825\std{.040} & 3.176\std{.086} & 4.035\std{.108} & \\
 
    \bottomrule
  \end{tabular}\end{adjustbox}
\end{table}

\subsection{Transfer across domains}\label{app:xdomain}
Here the programs cross between language and puzzles, with matched pretraining and no further search.
Table~\ref{tab:xfer-p2l} pretrains on the program optimized for each puzzle (Section~\ref{sec:exp-planning}) and
fine-tunes on the four language datasets; Table~\ref{tab:xfer-l2p} pretrains on the program optimized for
WikiText-103 and fine-tunes on the four puzzles, with the same training seeds as in Table~\ref{tab:planning}. Programs optimized
on Blocksworld and Sokoban are level with programs optimized on the language datasets themselves and ahead of the
hand-written program in 9 and 12 of 12 runs; the 8-puzzle and Countdown programs transfer less well. In the other
direction the WikiText-103 program solves more problems than the hand-written program on every puzzle, but no single
difference is significant, and it stays slightly below the program optimized on each task.
\begin{table}[H]
  \caption{Puzzle programs on language: test perplexity after a 10k-step fine-tune, mean\,$\pm$\,SD over 3 seeds
  (matched pretraining). Last two columns: pooled difference over the four datasets (12 seed-paired runs;
  positive: the puzzle program lower), $p$ in parentheses. First row: the program optimized on each language
  dataset itself, with the same three seeds (a subset of the nine runs of Table~\ref{tab:lang}).}
  \label{tab:xfer-p2l}
  \centering\small\setlength{\tabcolsep}{3pt}
  \begin{adjustbox}{max width=\linewidth}\begin{tabular}{@{}lcccccc@{}}
    \toprule
    Pretraining & WikiText-103 & TinyStories & CodeParrot & CodeSearchNet & vs hand-written & vs optimized on target \\
    \midrule
    \rowcolor{hl}Optimized on the target & 5.281\std{.162} & 2.795\std{.111} & 3.199\std{.099} & 3.976\std{.243} & &  \\
Blocksworld program & 5.259\std{.346} & 2.797\std{.152} & 3.151\std{.218} & 3.976\std{.287} & $+$0.280 {\scriptsize(.052)} & $+$0.017 {\scriptsize(.687)} \\
8-puzzle program & 5.434\std{.105} & 2.866\std{.062} & 3.253\std{.109} & 4.406\std{.451} & $+$0.086 {\scriptsize(.561)} & $-$0.177 {\scriptsize(.107)} \\
Sokoban program & 5.270\std{.151} & 2.782\std{.060} & 3.200\std{.051} & 4.011\std{.163} & $+$0.260 {\scriptsize(.082)} & $-$0.003 {\scriptsize(.960)} \\
Countdown program & 5.390\std{.126} & 2.862\std{.068} & 3.246\std{.033} & 4.553\std{.878} & $+$0.063 {\scriptsize(.728)} & $-$0.200 {\scriptsize(.224)} \\
Hand-written ($\pi_0$) & 5.458\std{.036} & 2.874\std{.030} & 3.300\std{.023} & 4.671\std{.770} & &  \\
No pretraining & 5.585\std{.055} & 2.939\std{.077} & 3.205\std{.004} & 5.676\std{.149} & &  \\
 
    \bottomrule
  \end{tabular}\end{adjustbox}
\end{table}
\begin{table}[H]
  \caption{The WikiText-103 program on the puzzles: test solve rate after a 10k-step fine-tune, mean\,$\pm$\,SD
  over 9 runs (Blocksworld) or 6 (others), matched pretraining; the other rows are those of
  Table~\ref{tab:planning}. Bottom rows: differences paired by training seed (positive: the WikiText-103 program
  solves more), $p$ in parentheses.}
  \label{tab:xfer-l2p}
  \centering\small\setlength{\tabcolsep}{3pt}
  \begin{adjustbox}{max width=\linewidth}\begin{tabular}{@{}lcccc@{}}
    \toprule
    Pretraining & Blocksworld (9--12 blocks) & 8-puzzle & Sokoban & Countdown \\
    \midrule
    Optimized on the task & 0.216\std{.254} & 0.246\std{.028} & 0.086\std{.030} & 0.161\std{.012} \\
Optimized on WikiText-103 & 0.158\std{.237} & 0.207\std{.136} & 0.070\std{.046} & 0.147\std{.014} \\
Hand-written ($\pi_0$) & 0.019\std{.040} & 0.172\std{.035} & 0.019\std{.013} & 0.136\std{.019} \\
No pretraining & 0.000\std{.000} & 0.014\std{.011} & 0.007\std{.003} & 0.056\std{.039} \\
\midrule
WikiText-103 program $-$ hand-written & $+$0.139 {\scriptsize(.101)} & $+$0.035 {\scriptsize(.590)} & $+$0.050 {\scriptsize(.052)} & $+$0.012 {\scriptsize(.371)} \\
WikiText-103 program $-$ optimized on the task & $-$0.057 {\scriptsize(.312)} & $-$0.038 {\scriptsize(.532)} & $-$0.016 {\scriptsize(.598)} & $-$0.013 {\scriptsize(.233)} \\
 
    \bottomrule
  \end{tabular}\end{adjustbox}
\end{table}

\section{Model size and downstream data size}\label{app:sweeps}
Both sweeps use WikiText-103 with raw text and evaluate the network each method's ten-round loop produced (as in
Appendix~\ref{app:loop}, not matched pretraining). Every model size runs the same search schedule on 2{,}000
synthetic streams per round.

\begin{figure}[H]
  \centering
  \includegraphics[width=0.55\textwidth]{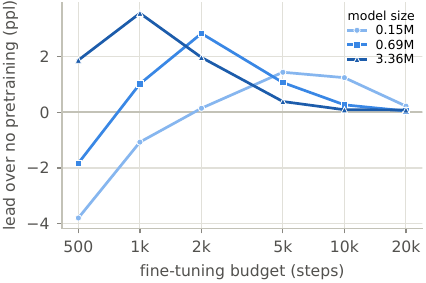}
  \caption{Head start by model size on WikiText-103: perplexity of a network with no synthetic pretraining minus
  \method{}'s, at each fine-tuning budget (best validation perplexity reached within the budget, mean over 3
  seeds; above 0: \method{} is ahead). Larger models reach a larger lead, earlier, and close the gap sooner. The
  two smallest pretrained networks start behind (at 500 steps, and also at 1k for 0.15M).}
  \label{fig:headstart}
\end{figure}

\textbf{Model size.} Table~\ref{tab:sweep-size} repeats the WikiText-103 comparison at 0.15M, 0.69M and 3.36M
parameters, and Figure~\ref{fig:headstart} follows the lead over no pretraining across fine-tuning budgets. The
head start is larger for larger models but comes earlier: at its peak a network with no synthetic pretraining
trails \method{} by 1.44 perplexity at 0.15M (after 5k fine-tuning steps), 2.83 at 0.69M (after 2k) and 3.56
at 3.36M (after 1k). A larger model also closes the gap sooner, so at the 10k-step budget of
Table~\ref{tab:sweep-size} the lead shrinks with size: over the hand-written curriculum it falls from 1.445 to
0.196 and 0.098 perplexity, and over no pretraining from 1.248 to 0.273 and 0.195. With three seeds none of
these differences is significant on its own ($p$ between .076 and .351). The synthetic budget is the same for
every size, so larger models receive less synthetic data per parameter; whether a curriculum scaled with the
model keeps the lead is untested.

\begin{table}[H]
  \caption{Model size sweep, $N{=}10$k WikiText-103 documents: test perplexity after a 10k-step fine-tune,
  mean\,$\pm$\,SD over 3 seeds. Last two columns: seed-paired differences (positive: \method{} lower), $p$ in
  parentheses.}
  \label{tab:sweep-size}
  \centering\small\setlength{\tabcolsep}{3.5pt}
  \begin{adjustbox}{max width=\linewidth}\begin{tabular}{@{}lccccc@{}}
    \toprule
    Parameters & \method{} & Hand-written ($\pi_0$) & No pretraining & \method{} $-$ hand & \method{} $-$ none \\
    \midrule
    0.15M & 8.254\std{.746} & 9.699\std{.885} & 9.502\std{.069} & $+$1.445 {\scriptsize(.093)} & $+$1.248 {\scriptsize(.110)} \\
0.69M & 5.312\std{.146} & 5.508\std{.082} & 5.585\std{.055} & $+$0.196 {\scriptsize(.112)} & $+$0.273 {\scriptsize(.076)} \\
3.36M & 4.063\std{.126} & 4.161\std{.097} & 4.258\std{.038} & $+$0.098 {\scriptsize(.351)} & $+$0.195 {\scriptsize(.110)} \\
 
    \bottomrule
  \end{tabular}\end{adjustbox}
\end{table}

\textbf{Downstream data.} Table~\ref{tab:sweep-n} varies the number $N$ of fine-tuning documents for the 0.69M
model. The lead is largest with the least data: at $N{=}2$k, \method{} is ahead of the hand-written curriculum
by 0.213 perplexity ($p{=}.040$) and of no pretraining by 0.410 ($p{=}.024$), the only significant differences
in the sweep. At larger $N$ the lead is smaller and does not change monotonically with $N$ (0.041, 0.196 and
0.053 over the hand-written curriculum at 5k, 10k and 20k), so we claim no trend beyond the smallest $N$.

\begin{table}[H]
  \caption{Downstream data sweep on WikiText-103, 0.69M model: test perplexity after a 10k-step fine-tune,
  mean\,$\pm$\,SD over 3 seeds. $N$ is the number of fine-tuning documents; $N{=}10$k is the main experiment
  (Table~\ref{tab:lang-loop}). Last two columns: seed-paired differences (positive: \method{} lower), $p$ in
  parentheses; bold: $p<.05$.}
  \label{tab:sweep-n}
  \centering\small\setlength{\tabcolsep}{3.5pt}
  \begin{adjustbox}{max width=\linewidth}\begin{tabular}{@{}lccccc@{}}
    \toprule
    $N$ & \method{} & Hand-written ($\pi_0$) & No pretraining & \method{} $-$ hand & \method{} $-$ none \\
    \midrule
    2{,}000 & 5.454\std{.058} & 5.667\std{.033} & 5.863\std{.066} & \textbf{$+$0.213} {\scriptsize(.040)} & \textbf{$+$0.410} {\scriptsize(.024)} \\
5{,}000 & 5.531\std{.056} & 5.572\std{.080} & 5.647\std{.016} & $+$0.041 {\scriptsize(.650)} & $+$0.116 {\scriptsize(.102)} \\
10{,}000 & 5.312\std{.146} & 5.508\std{.082} & 5.585\std{.055} & $+$0.196 {\scriptsize(.112)} & $+$0.273 {\scriptsize(.076)} \\
20{,}000 & 5.450\std{.085} & 5.503\std{.040} & 5.587\std{.022} & $+$0.053 {\scriptsize(.460)} & $+$0.137 {\scriptsize(.157)} \\
 
    \bottomrule
  \end{tabular}\end{adjustbox}
\end{table}

\section{Pretraining loss during the search}\label{app:dynamics}
\begin{figure}[H]
  \centering
  \includegraphics[width=\textwidth]{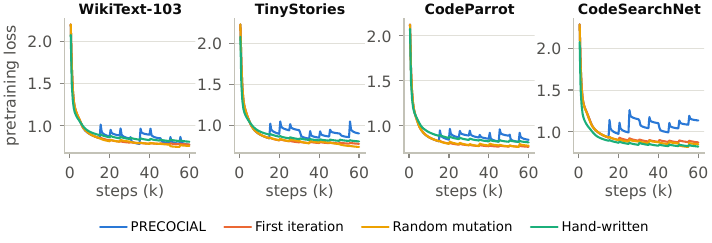}
  \caption{Pretraining loss during the search, concatenated over rounds (mean over 3 seeds); the $x$ axis is
  thousands of pretraining steps. From round 1 on, \method{} sits \emph{above} the first-iteration and
  random-mutation baselines (in round 0 all three train on the same first-iteration program) and spikes at each round boundary,
  where the next candidate changes the data, yet reaches the best downstream perplexity
  (Table~\ref{tab:lang-loop}).}
  \label{fig:pretrainloss}
\end{figure}

Figure~\ref{fig:dynamics} (Section~\ref{sec:exp-lang}) shows what the search does round by round, and
Figure~\ref{fig:pretrainloss} what the network sees while it happens.

In Figure~\ref{fig:pretrainloss}, from round 1 on \method{}'s pretraining loss is higher than that of the
first-iteration and random-mutation baselines on all four datasets, and in the last round it is also higher than the
hand-written program's on three of them (on WikiText-103 the hand-written program ends with the highest pretraining
loss of any method), while \method{}'s downstream perplexity is the lowest everywhere. Part of the gap is mechanical:
the first-iteration and hand-written programs show the network the same distribution in every round, and random
mutation only reweights, drops or duplicates the first-iteration program's rules, whereas an LLM rewrite can add new kinds of
data. This is also what produces the sawtooth: \method{}'s loss rises at 32 of its 36 round boundaries. We
therefore report the pattern without claiming that the optimized data is intrinsically harder to fit; separating
``harder data'' from ``newer data'' would need a run that replays the optimized programs in a fixed order.

\section{Epiplexity of the programs}\label{app:epiplexity}
We estimate the epiplexity of a program's data with the prequential estimate of \citet{finzi2026epiplexity}. A
fresh 0.69M network, with the architecture and optimiser of our experiments, makes a single pass over 40{,}000
newly generated streams of the program (5{,}000 steps of 8 streams, each truncated to 512 tokens), so that no
stream is seen twice. Before each update we record the mean per-token loss $\ell_t$ on the batch; after the pass
we measure the loss $H$ on 500 held-out streams of the same program, which estimates its time-bounded entropy. The
epiplexity estimate is $S=\sum_t(\ell_t-H)$, the area between the loss curve and the final loss, in nats per token
summed over steps. Table~\ref{tab:epiplexity} gives $S$ and $H$ for the 26 programs of this paper. Across them,
more epiplexity goes with lower WikiText-103 test perplexity after matched pretraining and a 10k-step fine-tune
(Spearman $\rho=-0.64$, $p<.001$; the perplexity of the run whose training seed is the program's search seed, or
the mean of three runs for the hand-written and procedural programs), whereas $H$ does not predict perplexity
($\rho=-0.08$, $p=.69$). The optimized WikiText-103 programs carry more epiplexity than their first-iteration
programs in all three searches (1659 against 1590, 1459 against 1397 and 1396 against 1293), and every group of
LLM-written programs carries more on average than the hand-written and procedural programs. Epiplexity separates
these broad sources of programs better than it ranks the programs of one search: the program kept after round 5
has more epiplexity than the final program in two of the three WikiText-103 searches, although the final programs
reach lower perplexity (Table~\ref{tab:lang}).
\begin{table}[H]
  \caption{Epiplexity $S$ (nats per token summed over steps) and final held-out loss $H$ (nats per token) of each
  program's synthetic data, measured with a fresh 0.69M network making one pass over 40{,}000 fresh streams. Mean
  and range over the programs of a group; all programs except the last group come from the WikiText-103 searches.}
  \label{tab:epiplexity}
  \centering\small
  \begin{tabular}{@{}lcccc@{}}
    \toprule
    Programs & $n$ & Epiplexity $S$ & Range of $S$ & $H$ \\
    \midrule
    Hand-written ($\pi_0$) & 1 & 1115 & 1115 & 0.986 \\
Procedural & 1 & 843 & 843 & 0.859 \\
First iteration & 3 & 1427 & 1293--1590 & 0.981 \\
Random mutation & 3 & 1380 & 1285--1557 & 0.974 \\
\method{} after round 5 & 3 & 1579 & 1233--1798 & 0.990 \\
\method{} & 3 & 1505 & 1396--1659 & 0.943 \\
Optimized on a puzzle (4 tasks $\times$ 3) & 12 & 1456 & 1044--1884 & 1.039 \\
 
    \bottomrule
  \end{tabular}
\end{table}

\section{Details of the reasoning and problem-solving tasks}\label{app:planning}
\begin{table}[H]
  \caption{Reasoning and problem solving, matched pretraining: final test perplexity on the answer tokens, mean\,$\pm$\,SD (same runs
  as Table~\ref{tab:planning}). Bottom rows: seed-paired difference in perplexity, positive when \method{} is
  lower.}
  \label{tab:planning-ppl}
  \centering\small\setlength{\tabcolsep}{4pt}
  \begin{adjustbox}{max width=\linewidth}\begin{tabular}{@{}lcccc@{}}
    \toprule
    Pretraining & Blocksworld & 8-puzzle & Sokoban & Countdown \\
    \midrule
    \method{} & 1.059\std{.075} & 1.165\std{.009} & 1.440\std{.041} & 1.635\std{.016} \\
First iteration & 1.124\std{.056} & 1.193\std{.030} & 1.625\std{.038} & 1.638\std{.014} \\
Hand-written ($\pi_0$) & 1.119\std{.089} & 1.185\std{.013} & 1.566\std{.048} & 1.640\std{.016} \\
No pretraining & 1.265\std{.075} & 1.285\std{.019} & 1.628\std{.023} & 1.843\std{.119} \\
\midrule
\method{} $-$ hand-written (ppl, $+$ = \method{} lower) & $+$0.061 {\scriptsize(.100)} & \textbf{$+$0.019} {\scriptsize(.009)} & \textbf{$+$0.127} {\scriptsize(.003)} & $+$0.005 {\scriptsize(.597)} \\
\method{} $-$ first iteration (ppl, $+$ = \method{} lower) & $+$0.065 {\scriptsize(.071)} & $+$0.027 {\scriptsize(.083)} & \textbf{$+$0.185} {\scriptsize(.001)} & $+$0.003 {\scriptsize(.733)} \\
 
    \bottomrule
  \end{tabular}\end{adjustbox}
\end{table}

\begin{figure}[H]
  \centering
  \includegraphics[width=\textwidth]{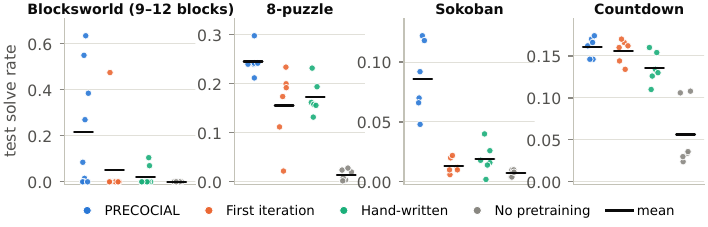}
  \caption{Every matched-pretraining run of Table~\ref{tab:planning} (dots; bars: means). On Blocksworld the
  solve rate is nearly all-or-nothing: it is a step function of the test perplexity the fine-tuned model reaches
  (Table~\ref{tab:bw-cliff}), so single runs sit at zero or well above it.}
  \label{fig:planning}
\end{figure}

\textbf{Blocksworld is nearly all-or-nothing.} A plan is about 30 actions (31 on average for our test problems) and every action must be legal
and correct, so a model either produces whole correct plans often or almost never. Across the 33 Blocksworld runs of
Table~\ref{tab:planning} the solve rate is a step function of the test perplexity the fine-tuned model reaches
(Table~\ref{tab:bw-cliff}; Spearman $\rho$ = $-0.78$ over 33 runs, $p = 6.5\times 10^{-8}$
\unskip): all 9 runs below 1.04 solve
some problems (0.015 to 0.635) and all 24 runs above it solve none, with no run in between. This is why single
Blocksworld runs sit either at zero or far above it (Figure~\ref{fig:planning}), and why we use 9 runs per method
there. Table~\ref{tab:planning-ppl} gives the perplexities behind those solve rates.
\begin{table}[H]
  \caption{Blocksworld: mean solve rate of the 33 runs of Table~\ref{tab:planning}, grouped by final test perplexity.}
  \label{tab:bw-cliff}
  \centering\small
  \begin{adjustbox}{max width=\linewidth}\begin{tabular}{@{}lccccc@{}}
    \toprule
    Test perplexity & $<$1.01 & 1.01--1.03 & 1.03--1.05 & $\geq$1.05 \\ \midrule
Runs & 5 & 2 & 2 & 24 \\
Mean solve rate & 0.463 & 0.095 & 0.043 & 0.000 \\
 
    \bottomrule
  \end{tabular}\end{adjustbox}
\end{table}

\section{The plans of Figure~\ref{fig:rollouts}}\label{app:rollouts}
\begin{figure}[H]
  \centering
  \includegraphics[width=0.86\textwidth]{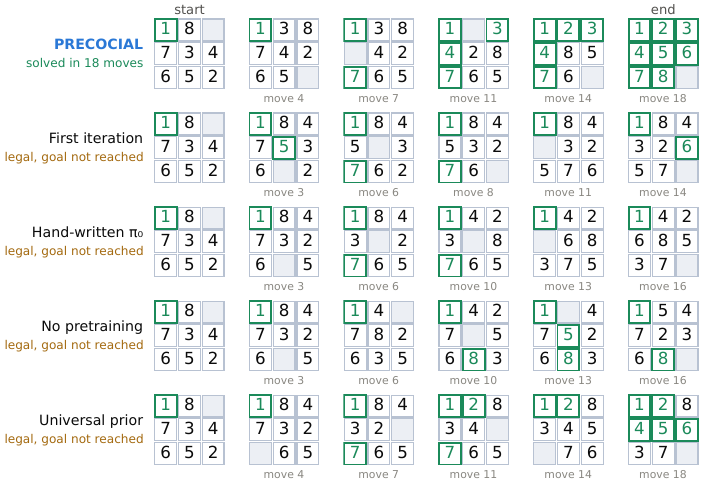}
  \caption{The 8-puzzle of Figure~\ref{fig:rollouts}: six evenly spaced states of each network's plan.}
  \label{fig:rollout-eight}
\end{figure}

\begin{figure}[H]
  \centering
  \includegraphics[width=0.92\textwidth]{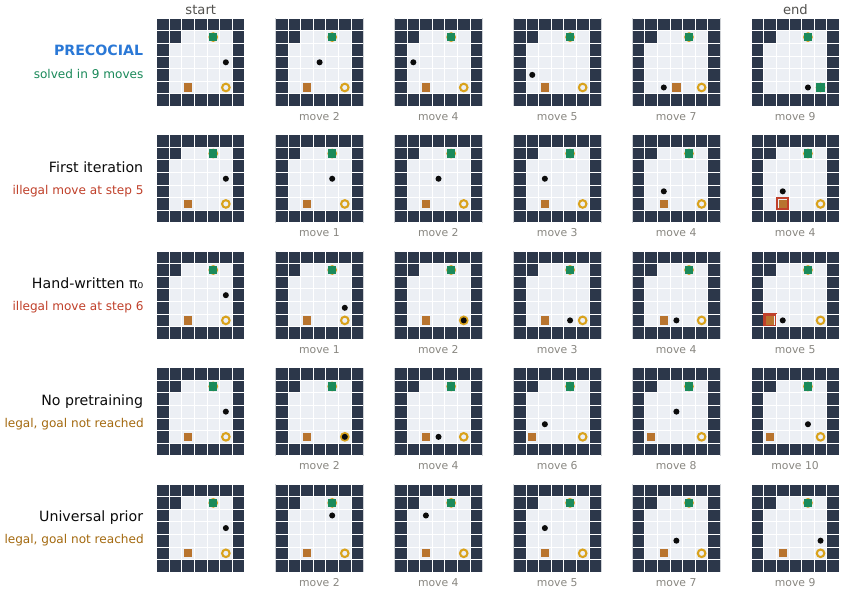}
  \caption{The Sokoban puzzle of Figure~\ref{fig:rollouts}: six evenly spaced states of each network's plan. The dot
  is the player; a red square marks the cell the illegal move tries to enter.}
  \label{fig:rollout-soko}
\end{figure}

\begin{figure}[H]
  \centering
  \includegraphics[width=\textwidth]{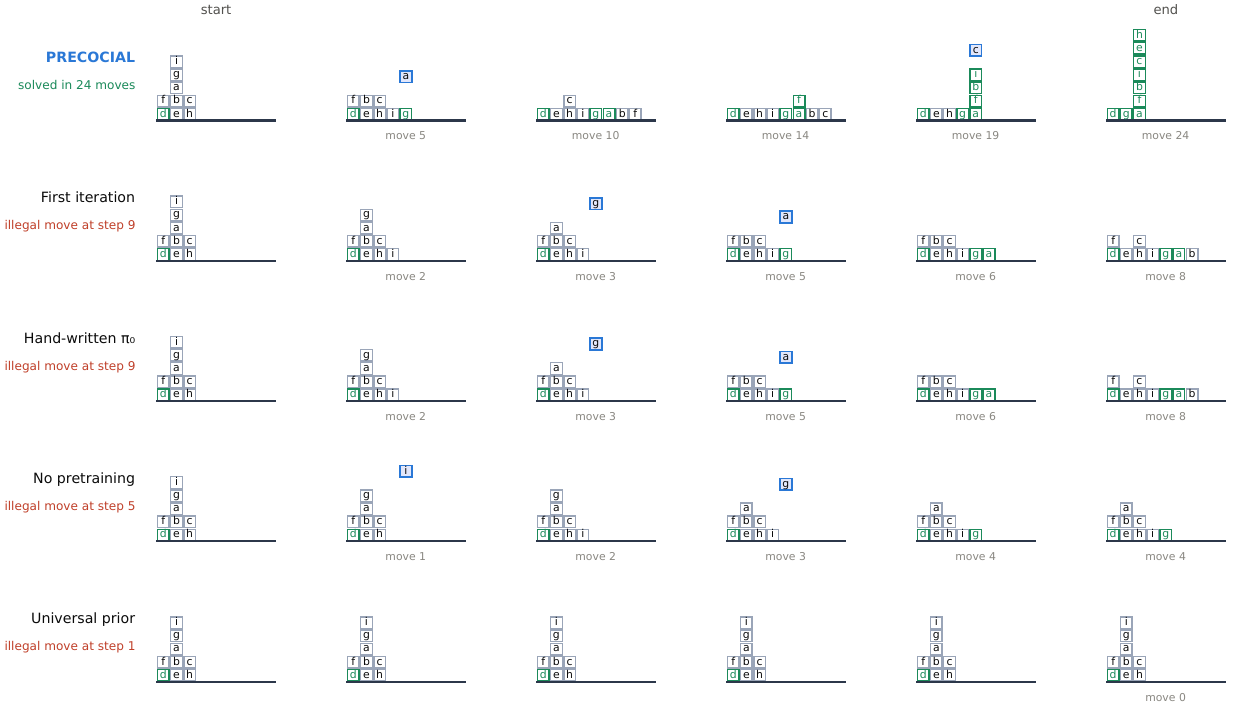}
  \caption{The Blocksworld puzzle of Figure~\ref{fig:rollouts} (9 blocks): six evenly spaced states of each
  network's plan; the block above the stacks is the one in the hand.}
  \label{fig:rollout-bw}
\end{figure}

The networks of Figure~\ref{fig:rollouts} were trained for the figure with the protocol of Table~\ref{tab:planning}:
each method's matched-pretrained network is fine-tuned on the task with training seed 64, and its greedy plan for
every test puzzle is replayed by the task's exact checker. Besides the methods of Table~\ref{tab:planning}, the
figure shows a \emph{universal prior}: a network pretrained with the same model and budget on the outputs of random
programs for a Brainfuck-like universal machine, our reimplementation after \citet{graumoya2024universal}. The
puzzle shown for each task is chosen by a fixed rule, not by hand: among the test puzzles that the \method{} network
solves (119 of 500 for the 8-puzzle, 43 of 500 for Sokoban and 82 of 200 for Blocksworld), keep those solved by the
fewest of the other networks trained for the figure, and take the first in test-set order. The other networks are
those shown and four more pretrained on reimplementations of published synthetic data (procedural generators, Dyck
languages, cellular automata and LIME-style tasks), and for Blocksworld also random mutation; for every task the
fewest is zero, and 38, 21 and 20 puzzles qualify. Figures~\ref{fig:rollout-eight}--\ref{fig:rollout-bw} show six
evenly spaced states of each plan. Table~\ref{tab:rollout-rates} gives the solve rates of these networks over three
training seeds; they are separate runs from those of Table~\ref{tab:planning}.
\begin{table}[H]
  \caption{Test solve rate of the networks trained for Figure~\ref{fig:rollouts}, mean\,$\pm$\,SD over training
  seeds 64, 65 and 66 (the figure shows seed 64).}
  \label{tab:rollout-rates}
  \centering\small
  \begin{tabular}{@{}lccc@{}}
    \toprule
    Pretraining & 8-puzzle & Sokoban & Blocksworld \\
    \midrule
    \rowcolor{hl}\method{} & 0.237\std{.003} & 0.073\std{.014} & 0.137\std{.237} \\
First iteration & 0.143\std{.120} & 0.014\std{.014} & 0.007\std{.012} \\
Hand-written ($\pi_0$) & 0.192\std{.074} & 0.037\std{.026} & 0.000\std{.000} \\
No pretraining & 0.021\std{.014} & 0.010\std{.002} & 0.000\std{.000} \\
Universal prior & 0.154\std{.053} & 0.049\std{.006} & 0.000\std{.000} \\
 
    \bottomrule
  \end{tabular}
\end{table}

\section{Cost}\label{app:cost}
\begin{table}[H]
  \caption{Cost of one search with ten candidates on the 0.69M model: WikiText-103 and CodeSearchNet, 3 seeds each
  (6 searches per row), all on RTX A6000 GPUs. The five time columns are GPU-hours of one search and add up to the
  total (up to rounding). How to read each column is explained below the table.}
  \label{tab:cost}
  \centering\footnotesize\setlength{\tabcolsep}{2.6pt}
  \begin{adjustbox}{max width=\linewidth}\begin{tabular}{@{}lrrrrrrrrr@{}}
    \toprule
    & Pretraining & \multicolumn{5}{c}{GPU-hours per search} & LLM & API & Test \\
    \cmidrule(lr){3-7}
    Search & steps & Pretrain & Fine-tune & LLM & Other & Total & calls & cost & ppl \\
    \midrule
    \method{} (one network) & 60,000 & 0.14 & 0.37 & 0.17 & 0.02 & 0.71\std{.12} & 10.8 & \$1.20 & 4.659 \\
FunSearch (network per candidate) & 150,000 & 0.36 & 0.37 & 0.12 & 0.02 & 0.88\std{.17} & 9.0 & \$1.10 & 4.890 \\
AlphaEvolve (network per candidate) & 150,000 & 0.35 & 0.36 & 0.12 & 0.02 & 0.85\std{.16} & 9.3 & \$1.08 & 4.955 \\
 
    \bottomrule
  \end{tabular}\end{adjustbox}
\end{table}
\textbf{Reading Table~\ref{tab:cost}.} Each row is one way of running a search that evaluates ten candidate
programs. \emph{Pretraining steps} is the number of synthetic pretraining steps summed over the whole search, not a
fine-tuning budget: \method{} pretrains one network for 15k steps for the first candidate and continues it for 5k
steps for each of the nine later ones ($15\text{k}+9{\times}5\text{k}=60\text{k}$), whereas FunSearch and
AlphaEvolve as published pretrain a fresh network for 15k steps for each of the ten candidates
($10{\times}15\text{k}=150\text{k}$). The five time columns split one search into synthetic pretraining;
fine-tuning the ten candidates for up to 10k steps each, which is identical under both protocols; waiting for the
LLM; everything else (generating the synthetic data, evaluation and bookkeeping); and their total, mean\,$\pm$\,SD
over the six searches. \emph{LLM calls} counts every call, including retries of programs that failed validation,
and \emph{API cost} prices them at Claude Sonnet~4 rates. \emph{Test ppl} is the test perplexity of the program
each search kept after a 10k-step fine-tune, averaged over the six searches (three on each dataset). One network
cuts pretraining time by 59--61\%, but fine-tuning is unchanged and takes 42--53\% of a search, so the total
falls by 17.1--19.5\%.

\textbf{Longer searches and larger models (projection).} From the WikiText-103 searches of Appendix~\ref{app:sweeps}
(3 per model size, RTX A6000), one candidate costs 0.039\,GPU-h of fine-tuning at 0.69M parameters and 0.067 at
3.36M, one LLM call 0.015, and everything else 0.003, while synthetic pretraining costs 2.54\,GPU-h (0.69M) and
5.07 (3.36M) per million steps. With one network a search of $n$ candidates pretrains $15\text{k}+5\text{k}\,(n{-}1)$
steps; with a fresh network per candidate, $15\text{k}\,n$. Table~\ref{tab:projection} multiplies these rates out.
These are projections, not runs. At ten candidates they reproduce the measured one-network searches (0.73 against
0.71\,GPU-h at 0.69M, 1.15 against 1.18 at 3.36M), but they predict a 24\% saving at 0.69M against the 17--20\%
measured in Table~\ref{tab:cost}, so they overstate the saving somewhat. The saving grows with the number of
candidates and with model size, because pretraining is a larger share of each candidate's cost for a larger model.
\begin{table}[H]
  \caption{Projected GPU-hours of one search, with one network continued across candidates (``one'') and with a
  fresh network per candidate (``fresh''), from the per-candidate rates above. Last column: synthetic pretraining
  steps summed over the search, one vs fresh.}
  \label{tab:projection}
  \centering\footnotesize\setlength{\tabcolsep}{4pt}
  \begin{adjustbox}{max width=\linewidth}\begin{tabular}{@{}rrrrrrrl@{}}
    \toprule
    & \multicolumn{3}{c}{0.69M model} & \multicolumn{3}{c}{3.36M model} & \\
    \cmidrule(lr){2-4}\cmidrule(lr){5-7}
    Candidates & one & fresh & saving & one & fresh & saving & Pretraining steps (one vs fresh) \\
    \midrule
    10 & 0.73 & 0.95 & 24\% & 1.15 & 1.61 & 28\% & 60,000 vs 150,000 \\
100 & 7.03 & 9.54 & 26\% & 11.1 & 16.1 & 31\% & 510,000 vs 1,500,000 \\
1,000 & 70.0 & 95.4 & 27\% & 110 & 161 & 31\% & 5,010,000 vs 15,000,000 \\
 
    \bottomrule
  \end{tabular}\end{adjustbox}
\end{table}

\section{Rule classification}\label{app:rules}
Rules are classified by keyword matching on their names, which are the LLM's own and usually say what the rule
does: \emph{planning} (e.g.\ \texttt{pddl}, \texttt{plan}, \texttt{action}, \texttt{state}, \texttt{precond},
\texttt{grid}, \texttt{move}, \texttt{puzzle}, \texttt{sokoban}), \emph{arithmetic} (\texttt{arith}, \texttt{expr},
\texttt{math}, \texttt{fraction}, \texttt{make24}), \emph{code syntax} (\texttt{func}, \texttt{class\_def},
\texttt{import}, \texttt{method}, \texttt{param}, \texttt{try\_catch}), list and string operations, logic, cellular
automata, and formal structure (\texttt{nest}, \texttt{dyck}, \texttt{tree}, \texttt{cfg}, \texttt{stack}); a name takes the
first matching type in a fixed order. The share of a type is its fraction of the synthetic data generated in a
round, by volume. The classifier is a keyword map,
so single percentages are approximate; the script that assigns every name is released with the code.

\begin{table}[H]
  \begin{minipage}[b]{0.52\textwidth}
    \centering\footnotesize\setlength{\tabcolsep}{2pt}
    \begin{adjustbox}{max width=\linewidth}\begin{tabular}{@{}lcccc@{}}
      \toprule
      Task & $\pi_0$ & First iter. & All cand. & Kept \\
      \midrule
      Blocksworld & 0.0 & 0.0\std{0.0} & 4.3\std{6.7} & 3.3\std{4.9} \\
8-puzzle & 0.0 & 8.4\std{14.5} & 18.1\std{15.1} & 22.5\std{10.5} \\
Sokoban & 0.0 & 0.0\std{0.0} & 9.4\std{7.3} & 12.3\std{9.2} \\
Countdown & 3.6 & 3.4\std{2.2} & 20.3\std{10.8} & 22.8\std{6.7} \\
CodeParrot & 0.0 & 0.0\std{0.0} & 8.0\std{12.6} & 20.1\std{20.1} \\
CodeSearchNet & 0.0 & 0.0\std{0.0} & 9.4\std{12.3} & 17.2\std{18.5} \\
 
      \bottomrule
    \end{tabular}\end{adjustbox}
    \captionof{table}{Share of synthetic data (\%) from rules of the task-matching type (Figure~\ref{fig:rules}),
    mean\,$\pm$\,SD over three searches: in $\pi_0$, the first-iteration program, all candidates of rounds 0--9,
    and the kept program.}
    \label{tab:rules}
  \end{minipage}\hfill
  \begin{minipage}[b]{0.45\textwidth}
    \centering
    \includegraphics[width=\textwidth]{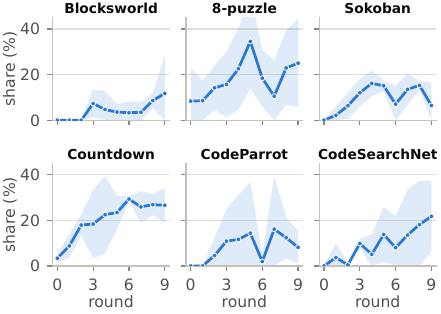}
    \captionof{figure}{Task-matching share by search round, mean\,$\pm$\,SD over three searches: planning rules
    for Blocksworld, the 8-puzzle and Sokoban, arithmetic for Countdown, code syntax for code.}
    \label{fig:rules}
  \end{minipage}
\end{table}

\end{document}